\documentclass[10pt,twocolumn,letterpaper]{article}

\usepackage{cvpr}              

\definecolor{cvprblue}{rgb}{0.21,0.49,0.74}
\usepackage[pagebackref,breaklinks,colorlinks,allcolors=cvprblue]{hyperref}
\usepackage{colortbl}
\usepackage{threeparttable}
\usepackage{stfloats}
\usepackage{float}

\definecolor{First}{rgb}{0.95, 0.62, 0.61}
\definecolor{Second}{rgb}{0.97,0.81,0.63}
\definecolor{Third}{rgb}{1.0, 0.97, 0.70}

\usepackage[T1]{fontenc}

\usepackage{styles}
\usepackage{multirow}
\usepackage{subcaption}
\usepackage{pgfplots}
\usepackage{pgfplotstable}
\usepackage{xcolor}
\usepackage{graphicx, overpic}
\usepackage{nicematrix}
\usepackage{makecell}
\usepackage{siunitx}

\newcommand{\locationrow}[1]{
\begin{minipage}{0.35\textwidth}  
    \centering
    \begin{tikzpicture}
        \node [anchor=south west] (evi) at (0,0)
            {
                \includegraphics[width=0.92\linewidth, height=0.45\linewidth, trim=20 58 70 80, clip]{sec/images/temporal_sampling/#1_evi_curve.pdf}
            };
        
        \node [anchor=south west] at (0.12\linewidth, 0.3\linewidth)
            {
                \includegraphics[width=0.25\linewidth, height=0.15\linewidth, trim=0 25\% 0 25\%, clip]{sec/images/temporal_sampling/#1_map.pdf}
            };
    \end{tikzpicture}
\end{minipage}
\begin{minipage}{0.64\textwidth}  
    \centering
        \includegraphics[width=\linewidth]{sec/images/temporal_sampling/#1_image_lighter.pdf}
\end{minipage}
}

\def\paperID{EarthVision-90} 
\def\confName{EarthVision}
\def\confYear{2025}

\title{
\DATAx: A Global Seasonal Dataset for Geospatial Foundation Models in Ecology
}

\author{
    {Elena Plekhanova\textsuperscript{1}}\\
    {\tt\small elena.plekhanova@wsl.ch}
    \and
    {Damien Robert\textsuperscript{2}}\\
    {\tt\small damien.robert@uzh.ch}
    \and
    {Johannes Dollinger\textsuperscript{2}}\\
    {\tt\small johannes.dollinger@uzh.ch}
    \and
    {Emilia Arens\textsuperscript{2}}\\
    {\tt\small emilia.arens@uzh.ch}
    \and
    {Philipp Brun\textsuperscript{1}}\\
    {\tt\small philipp.brun@wsl.ch}
    \and
    {Jan Dirk Wegner\textsuperscript{2}}\\
    {\tt\small jandirk.wegner@uzh.ch}
    \and
    {Niklaus Zimmermann\textsuperscript{1}}\\
    {\tt\small niklaus.zimmermann@wsl.ch}
    \and
    {\textsuperscript{1}Land Change Science, Swiss Federal Research Institute WSL, Birmensdorf, Switzerland}\\
    {\textsuperscript{2} DM$3$L, University of Zurich, Zurich, Switzerland}\\
}

\begin{document}

\twocolumn[{%
    \renewcommand\twocolumn[1][]{#1}%
    \maketitle
    \vspace{-2.5em}
    \begin{center}
    \begin{minipage}{0.49\textwidth}
        \centering
        \includegraphics[width=\linewidth]{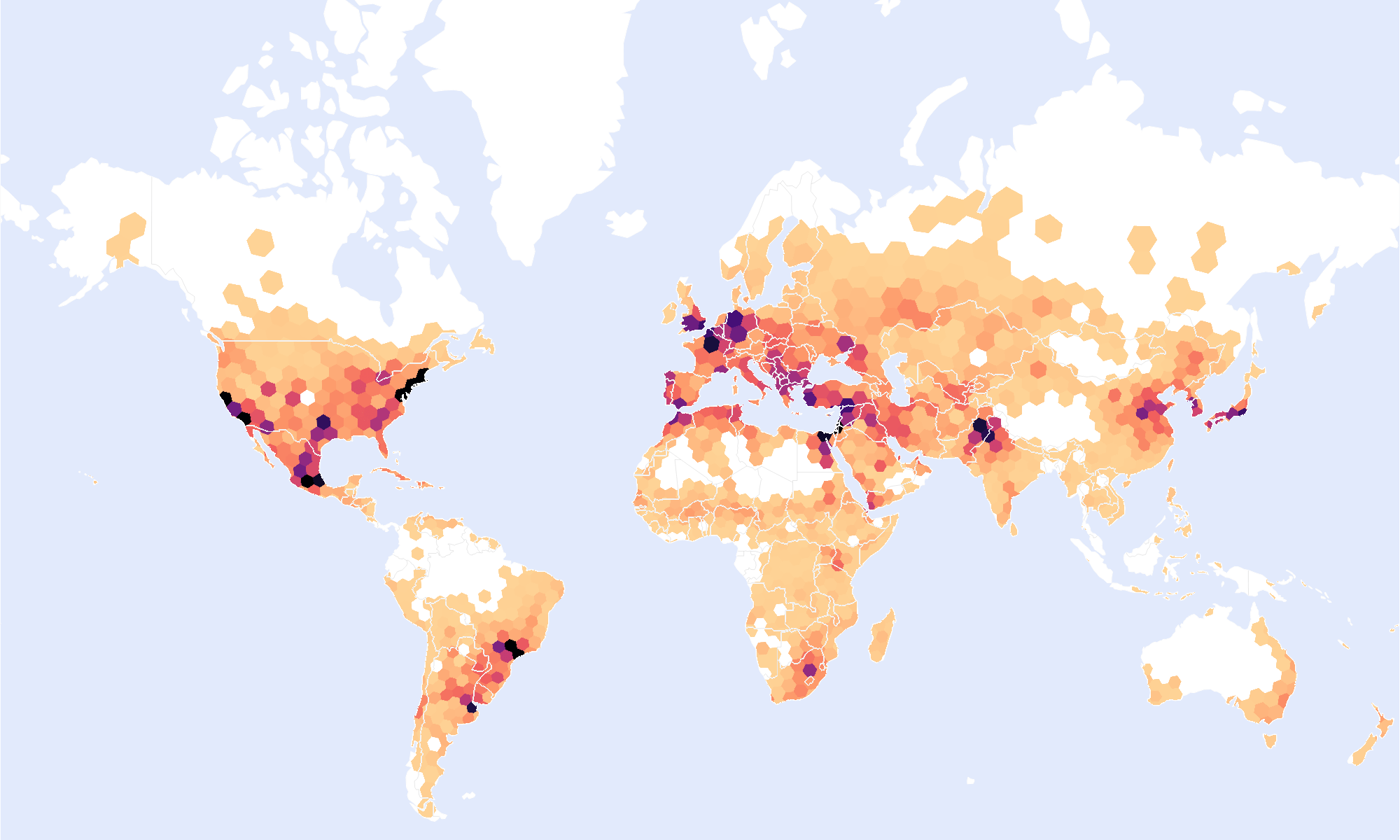}
        \\
        {\small (a) Spatial distribution of SSL4EO-S12~\cite{ssl4eo}}
        \label{fig:spatial_sampling_ssl4eo}
    \end{minipage}
    \hfill
    \begin{minipage}{0.49\textwidth}
        \centering
        \includegraphics[width=\linewidth]{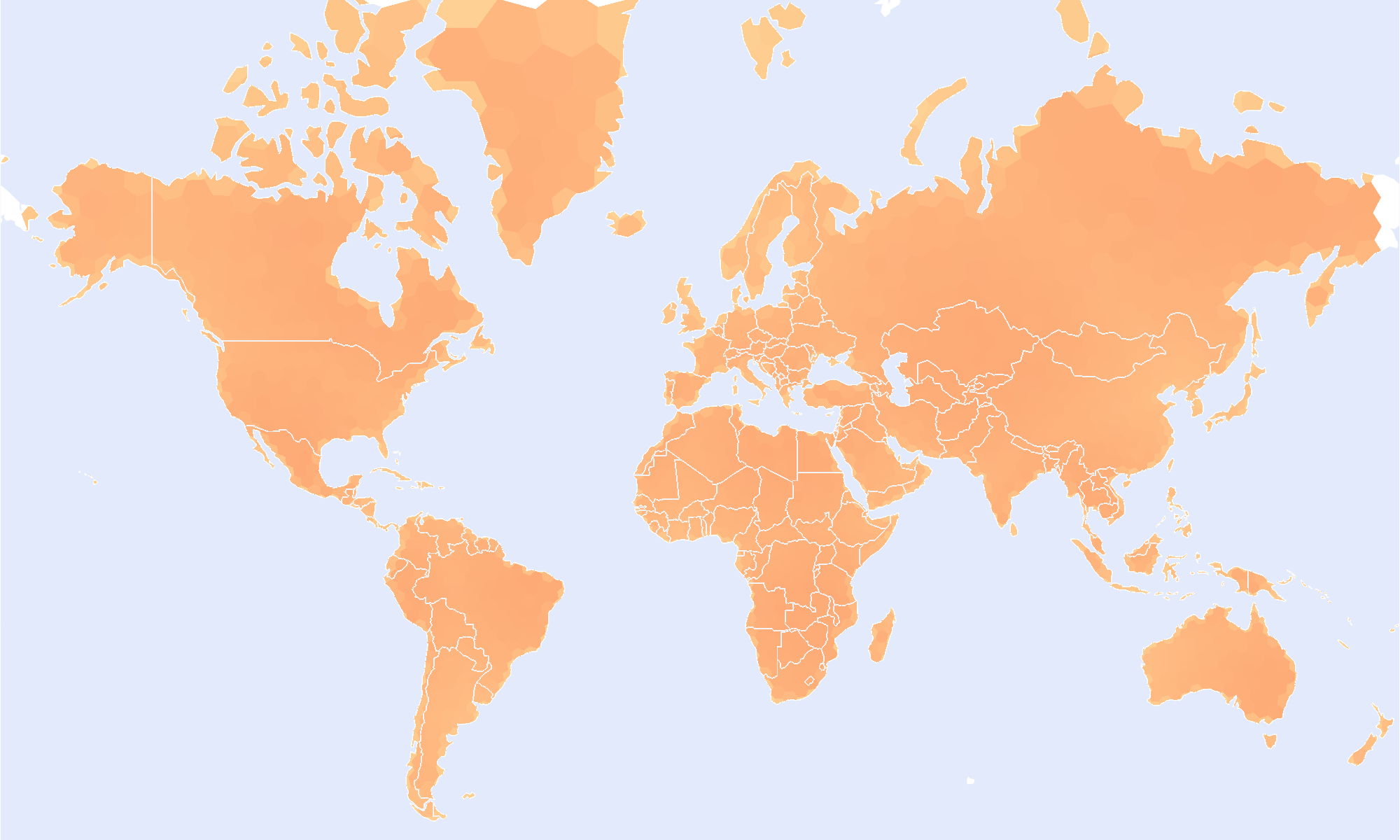}
        \\
        {\small (b) Spatial distribution of \DATA}
        \label{fig:spatial_sampling_ssl4eco}
    \end{minipage}

    \vspace{0.5em}

    \begin{minipage}{0.98\textwidth}
        \centering
        \includegraphics[width=\linewidth]{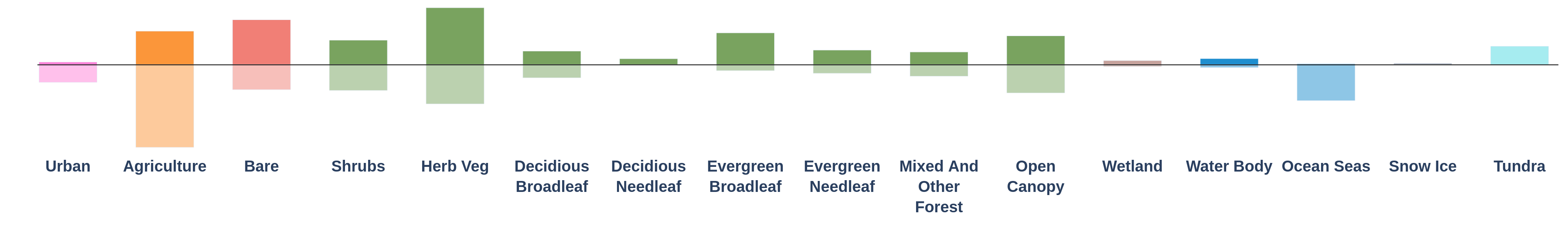}
        \\
        {\small (c) Copernicus land cover~\cite{landcover} distribution for \DATA (upwards) and SSL4EO-S12~\cite{ssl4eo} (downwards)}
        \label{fig:spatial_sampling_landcover}
    \end{minipage}

    
    \captionof{figure}{
        We propose \DATAx, a multi-date Sentinel-2 dataset for pretraining foundation models targeted for macroecological applications.
        Unlike comparable datasets (a), \DATA uniformly covers the entire landmass (b), thus capturing all environment types without favoring urban and agricultural areas, or ignoring entire ecoregions (c).
    }
    \label{fig:spatial_sampling}
\end{center}

}]

\begin{abstract}

\vspace{-1em}

With the exacerbation of the biodiversity and climate crises, macroecological pursuits such as global biodiversity mapping become more urgent. 
Remote sensing offers a wealth of Earth observation data for ecological studies, but the scarcity of labeled datasets remains a major challenge.
Recently, self-supervised learning has enabled learning representations from unlabeled data, triggering the development of pretrained geospatial models with generalizable features.
However, these models are often trained on datasets biased toward areas of high human activity, leaving entire ecological regions underrepresented.
Additionally, while some datasets attempt to address seasonality through multi-date imagery, they typically follow calendar seasons rather than local phenological cycles.
To better capture vegetation seasonality at a global scale, we propose a simple phenology-informed sampling strategy and introduce corresponding \DATAx, a multi-date Sentinel-2 dataset, on which we train an existing model with a season-contrastive objective.
We compare representations learned from \DATA against other datasets on diverse ecological downstream tasks and demonstrate that our straightforward sampling method consistently improves representation quality, highlighting the importance of dataset construction.
The model pretrained on \DATA reaches state of the art performance on 7 out of 8 downstream tasks spanning (multi-label) classification and regression.
We release our code, data, and model weights to support macroecological and computer vision research at~\\
{\vspace{-0.1em}\noindent\small\url{https://github.com/PlekhanovaElena/ssl4eco}}.

\end{abstract}    
\vspace{-1em}
\section{Introduction}
\label{sec:intro}

Biodiversity is essential for ecosystem stability and human well-being, yet it faces an unprecedented crisis due to habitat loss and climate change~\cite{ipbes}. 
Recognized as a global priority (SDG 15)~\cite{sgds}, biodiversity loss ranks among the most severe risks of the next decade~\cite{wef-risk}. 
This intensifying crisis calls for macroecological studies to understand spatiotemporal biodiversity patterns and identify priority areas for conservation~\cite{ipbes}.
Mapping changes in biodiversity, habitats, and land-use (\eg deforestation, urban or agricultural expansion) over time is essential for conservation planning~\cite{sdm4conservation, globalmaps}.
Central to these efforts is monitoring vegetation change, as vegetation forms the primary structure of most terrestrial ecosystems and shapes biodiversity patterns and ecosystem functions~\cite{turner2003remote}.


Remote sensing is a powerful tool for monitoring vegetation change at broad spatial and temporal scales~\cite{xie2008}. 
It provides consistent, repeated, global observations, enabling the detection of subtle shifts in vegetation health, species composition, and phenology---insights often unattainable through ground-based methods~\cite{fang2019, fischer2006}. 
Several open-access satellite products support vegetation monitoring, each with distinct strengths and limitations (see Appendix \ref{sec:details_dataset_construction}).
This work focuses on Sentinel-2 due to its widespread use for large-scale vegetation monitoring~\cite{lang2023high, sialelli2024agbd, karaman2025gsr4b}, but our conclusions remain applicable and may be extended to other satellite products.


To extract ecological insights from remote sensing data, initial approaches relied on handcrafted features and classical machine learning~\cite{gislason2006random, belgiu2016random}.
Deep learning has since revolutionized the field by automating feature extraction for tasks with annotated datasets~\cite{zhu2017deep, ma2019deep}.
Recently, self-supervised learning (SSL) has gained traction for learning rich representations from large, unlabeled datasets~\cite{gui2024}, with successful applications in the analysis of natural language~\cite{devlin2018bert}, natural images~\cite{oquab2023dinov2}, and remote sensing data~\cite{ayush2021geography}.
The resulting pretrained models produce representations that generalize to downstream tasks, making these so-called \textit{foundation models} (FMs) particularly suitable for applications where labeled data is scarce or costly, such as large-scale ecological studies~\cite{wang2022self}.

The size and diversity of the pretraining dataset largely influences the generalizability of the learned representations~\cite{radford, zhai}.
While research on geospatial FMs operating on georeferenced data (GFMs) is an active field of study, most effort is currently geared towards new model architectures and SSL pretraining tasks, and little attention is given to the design of pretraining datasets.
This oversight is critical, as the geographical distribution of training data significantly influences model performance~\cite{, roscher2024betternot, purohit2025does}. 
For biodiversity applications in particular, existing GFMs are often trained on datasets which fail to capture important spatiotemporal ecological patterns, as summarized in \tabref{tab:related_work_datasets}.
First, the geographic sampling is often biased towards human activity, hence over-representing urban and agricultural areas while neglecting entire biomes. 
Second, multi-temporal datasets are typically sampled following calendar seasons, failing to account for local phenological cycles, essential to biodiversity monitoring.

In this work, we propose a dataset construction recipe targeted towards the development of foundation models for ecology.
Specifically, we propose to sample locations uniformly across the landmass, rather than around large urban areas~\cite{seco, ssl4eo}, and sample dates based on local phenological cycles, rather than calendar seasons~\cite{seco, ssl4eo}.
Following this protocol, we introduce \DATAx, a pretraining dataset of multispectral, multi-date Sentinel-2 patches of $256 \times 256$ pixels, uniformly sampled across $250$k locations around the globe and capturing local phenology, as shown in \figref{fig:spatial_sampling} and \figref{fig:temporal_sampling}.
From \DATAx, we derive \SECOx, a seasonality-aware SeCo~\cite{seco} model, and compare its embeddings against off-the-shelf GFMs on diverse macroecological tasks.
We show that \SECO equals or exceeds the performance of all other baselines on 7 out of 8 downstream tasks spanning (multi-label) classification and regression, with larger gaps of $+2$ mAP on BigEarthNet-10\%~\cite{bigearthnet} and $+3$ to $+4$ R$^2$ in regression of climatic variables and biomass.


Far from claiming a new SSL training or backbone, this work stresses the importance of dataset design, and how a straightforward spatiotemporal sampling protocol may consistently benefit GFMs downstream applications.
We publicly release our datasets, code, and weights at \url{https://github.com/PlekhanovaElena/ssl4eco}, hoping to foster both downstream macroecological studies and methodological computer vision research with a concern for environmental applications.
The contributions of this work are as follows:

\begin{itemize}
  \item \DATAx: a novel multi-temporal Sentinel-2 pre-training dataset with uniform global distribution and vegetation phenology-based seasonal sampling.
  \item \SECOx: a seasonality-aware geospatial foundation model pretrained on \DATAx.
  \item New macroecological downstream tasks for benchmarking geospatial foundation models.
\end{itemize}


\section{Related Work}
\label{sec:related}


In this section, we provide an overview of existing remote sensing datasets used for pretraining geospatial foundation models, with a focus on their spatiotemporal distribution.
We then introduce several such foundation models relevant to this work.  

\begin{table}
    \small
    \centering
    
    \begin{NiceTabular}{@{}lcll@{}}
    \toprule    
    \multirow{2}{*}{Dataset} & \multicolumn{2}{c}{Locations} & \multirow{2}{*}{Seasons} \\
    & {\footnotesize Number} & {\footnotesize Distribution} & \\
    \midrule
    BigEarthNet~\cite{bigearthnet} & $600$k & Europe & -  \\
    SEN12MS~\cite{schmitt2019sen12ms} & $280$k & Around cities & Calendar  \\
    SeCo~\cite{seco} & $200$k & Around cities & Random \\
    S2-100k~\cite{satclip} & $100$k & Lat-lon uniform & - \\
    Planted~\cite{pazos2024planted} & $3.0$M & Semi global & - \\
    SatlasPretrain~\cite{saltas} & $3.0$M & Semi global & -  \\
    SSL4EO-S12~\cite{ssl4eo} & $250$k & Around cities & Calendar \\
    MajorTOM-Core~\cite{majortom} & $2.2$M & Global uniform & - \\
    \midrule
    \rowcolor{lightblue} \textbf{\DATA} & $250$k & Global uniform & EVI-based \\
    \bottomrule
    \end{NiceTabular}
   
    \caption{
        Comparison of the spatiotemporal sampling of popular 
        pretraining datasets for geospatial foundation models.
        Our sampling of \DATA is designed to fully capture both global geographic diversity and local climatic and phenological seasonality.
    }
    \label{tab:related_work_datasets}
\end{table}

\vspace{-1em}
\paragraph{Pretraining Remote Sensing Datasets.}
Numerous labeled datasets have been proposed to employ remote sensing imagery for mapping urban or agricultural landscapes~\cite{schneider2010mapping, garnot2021panoptic, garioud2023flair}. 
However, these generally do not offer the 
spatial and seasonal 
coverage necessary to macroecological research, as summarized in \tabref{tab:related_work_datasets}.
%
Existing unlabeled pretraining datasets for SSL models focus predominantly on regions experiencing high human impact, often neglecting areas crucial for ecological research and conservation.
For instance, SEN12MS~\cite{schmitt2019sen12ms}, SeCo~\cite{seco}, and SSL4EO-S12~\cite{ssl4eo} datasets are sampled around large cities, mainly encompassing urban and agricultural zones (\figref{fig:spatial_sampling}).
While BigEarthNet~\cite{bigearthnet} does sample diverse vegetation types, it only covers Europe. Other datasets such as SatlasPretrain~\cite{saltas},
S2-100K~\cite{satclip}, 
and Planted~\cite{pazos2024planted} have better geographic coverage, but with significant gaps in the tropics due to high cloud coverage and either undersample or ignore Arctic tundra entirely.
Yet, tropical rainforests harbor the highest levels of biodiversity on the globe~\cite{ipbes}, making their underrepresentation in training datasets problematic. 
Similarly, the Arctic region is central to many environmental processes such as the thawing of Arctic permafrost which introduces one of the greatest uncertainties in current climate models~\cite{schuur}. 
Interestingly, Major-TOM-Core~\cite{majortom} uniformly covers the entire landmass, but at a single date, failing to capture seasonality.
%
Despite the importance of seasonality for ecosystems and ecological research~\cite{pheno}, few datasets provide multi-temporal imagery at each location.
SeCo~\cite{seco} randomly selects $5$ dates across the year separated by approximately $3$ months.
SEN12MS~\cite{schmitt2019sen12ms} and SL4EO-S12~\cite{ssl4eo} select $4$ dates within seasonal windows defined based on calendar dates.
However, these sampling approaches treat all locations equally, resulting in datasets that overlook the reality of local climatic and ecological conditions.
Indeed, regions near the tropics may have longer leaf-on seasons, while desert or Arctic regions may see very brief events of vegetation activity with a large portion of the year being dry or snow-covered. 
Likewise, the beginning and end of dormancy periods may be shifted in the year, depending on local climatic conditions.
%
In this work, we propose a simple sampling strategy that fully covers the global diversity of landscapes (\figref{fig:spatial_sampling}) and local seasonality (\figref{fig:temporal_sampling}).
Our goal is to design datasets for learning representation better suited for downstream ecological applications. 

\begin{table}[t]
    \centering
    \small
    \begin{NiceTabular}{@{}llll@{}}
        \toprule
        Model & Dataset & Backbone & Pretraining \\
        \midrule
        SeCo~\cite{seco}~\cite{seco} & {\footnotesize SeCo~\cite{seco}} &  {\footnotesize ResNet50~\cite{resnet}}  & {\footnotesize SeCo~\cite{seco}} \\
        SatMAE~\cite{satmae} & {\footnotesize fMoW~\cite{fmow, satmae}} & {\footnotesize ViT-L~\cite{vit}} & {\footnotesize MAE~\cite{mae}}  \\
        Satlas~\cite{saltas} & {\footnotesize SatlasPretrain~\cite{saltas}} & {\footnotesize Swin-B~\cite{swin}} & {\footnotesize Supervised} \\
        Croma~\cite{croma} & {\footnotesize SSL4EO-S12~\cite{ssl4eo}} & {\footnotesize ViT-L~\cite{vit}} & {\footnotesize MAE~\cite{mae}} \\
        SSL4EO~\cite{ssl4eo} & {\footnotesize SSL4EO-S12~\cite{ssl4eo}}  & {\footnotesize ResNet50~\cite{resnet}}  & {\footnotesize MoCov2~\cite{mocov2}} \\
        DOFA~\cite{dofa} & {\footnotesize DOFA~\cite{dofa}} & {\footnotesize ViT-L~\cite{vit}}  & {\footnotesize MAE~\cite{mae}}   \\
        \arrayrulecolor{black}  
        \hline
        \rowcolor{lightblue} \textbf{\SECO} &{\footnotesize \textbf{\DATA}} & {\footnotesize ResNet50~\cite{resnet}} & {\footnotesize SeCo~\cite{seco}}  \\
        \bottomrule
    \end{NiceTabular}
    \caption{
        Overview of
        recent
        image-based
        geospatial foundation models.
        We focus on models trained to process Sentinel-2 data, for fair comparison with our pretraining setting.
        While we release a new pretrained image-based model \SECOx, our focus is \textit{not} on the design of a backbone or pretraining method, but on the impact of the pretraining dataset.
    }
    \label{tab:related_work_models}
\end{table}

\vspace{-1.5em}
\paragraph{Geospatial Foundation Models.}
Advances in self-supervised learning have recently allowed to learn generalizable representations from the wealth of public, unlabeled, satellite imagery~\cite{ayush2021geography, wang2022self}.
Masked image modeling~\cite{mae} methods typically leverage symmetries inherent to remote sensing data to reconstruct masked spectral bands~\cite{satmae}, time steps~\cite{dumeur2024self}, both~\cite{jakubik2023foundation}, or other modalities~\cite{tseng2023lightweight, croma, omnisat, mmearth}.
Alternatively, contrastive approaches~\cite{chen2020simple, moco, mocov2} learn to align latent representations of imagery from different seasons~\cite{seco, ssl4eo} or modalities~\cite{anysat}.
Another direction learns implicit geolocation representations by aligning spatial coordinates with terrestrial~\cite{geoclip} or satellite~\cite{satclip} imagery, or species occurrences~\cite{sinr}.
%
Moving beyond the focus on the design of self-supervised method or model architecture, our work sheds light on the importance of pretraining datasets. 
We use existing SSL methods to pretrain on our \DATA dataset and analyze resulting representations with available comparable image-based geospatial foundation models, as summarized in \tabref{tab:related_work_models}.
\section{Method}

We detail our proposed dataset construction approach in \secref{sec:method_dataset} and pretrained model in \secref{sec:method_model}.


\subsection{\DATA Dataset}
\label{sec:method_dataset}

\begin{figure*}[t]
    \centering


    \locationrow{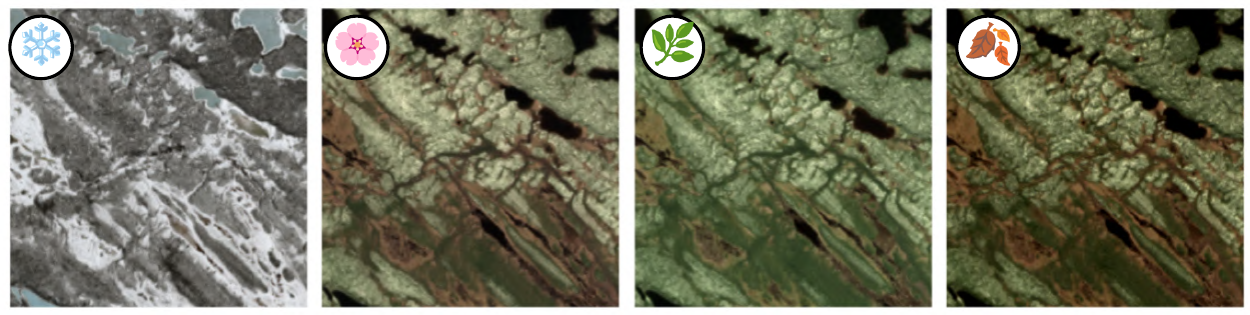}
    \locationrow{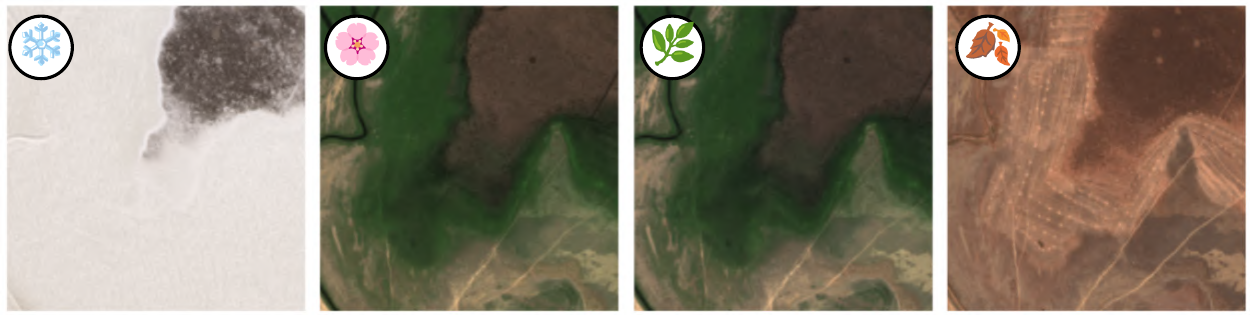}
    \locationrow{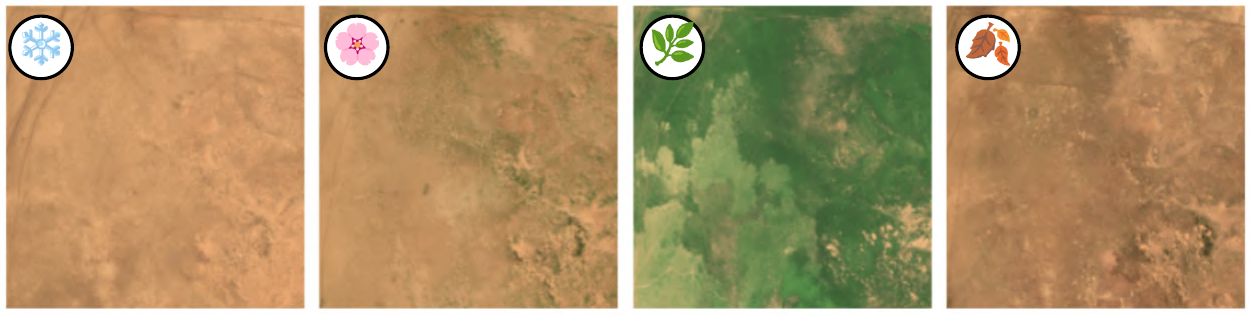}

    \vspace{0.5em}

    \begin{minipage}{0.35\textwidth}
        \centering
        \small
        (a) EVI-based seasons
    \end{minipage}
    \begin{minipage}{0.64\textwidth}
        \centering
        \small
        (b) Seasonal images
    \end{minipage}

    \caption{
        Unlike previous works which sample seasonal images based on calendar dates~\cite{schmitt2019sen12ms, seco, ssl4eo} (dashed lines in (a)), we define phenology-informed, local seasons based the Enhanced Vegetation Index~\cite{justice1998moderate, mcduserguide} (colored sections in (a)).
        As a result, our \DATA dataset covers the full cycle of vegetation activity at each location (b), capturing patterns otherwise missed by calendar sampling.
    }
    \label{fig:temporal_sampling}
\end{figure*}


Our dataset sampling recipe aims at capturing phenology-informed patterns anywhere on Earth.
For more details on our dataset construction protocol, please see \secref{sec:details_dataset_construction}.

\vspace{-1em}
\paragraph{Spatial Sampling.}
Similar to Major-TOM~\cite{majortom}, we uniformly sample geolocations across the globe using a regular grid, accounting for distortions long the latitude.
We only sample positions across the landmass, with a $23$ km spacing between points, yielding $250$k geolocations.
This sampling size is chosen to allow comparison with similar pretraining datasets~\cite{seco,ssl4eo}.
As shown in \figref{fig:spatial_sampling}, the resulting dataset follows the natural distribution of land use, without focusing on urban or agricultural areas.

\vspace{-1em}
\paragraph{Seasonal Sampling.}
Vegetation seasonality primarily depends on local temperature and light regimes, themselves primarily driven by latitude~\cite{partanen1998effects, chuine2004grape}, altitude~\cite{korner1999alpine, keller2003role}, and rainfall seasonality~\cite{dubois2014indonesian, cheng2021vegetation} (\eg monsoon regions, or Mediterranean and Savannah biomes).
To capture local seasonality, we sample $4$ dates at each location.
Unlike previous works which define seasons globally based on calendar dates~\cite{schmitt2019sen12ms, seco, ssl4eo}, we sample based on local plant phenology.
To this end, we use the Enhanced Vegetation Index (EVI) from the MCD12Q2 v6.1~\cite{modis} product of the MODIS~\cite{justice1998moderate} satellite mission.
For each location, we define the $4$ seasons spring, summer, autumn, and winter as intervals between the Greenup, Maturity, Senescence, and Dormancy variables (see \secrefshort{sec:evi_definition} for details).
By sampling a date in each of these phenological seasons, we aim to better seize the diversity of vegetation states at each location than calendar or random sampling, as illustrated in \figref{fig:temporal_sampling}. 

\vspace{-1em}
\paragraph{Modality.}
We apply our spatiotemporal sampling strategy to create \DATAx, a global, multi-temporal dataset of satellite imagery.
We choose to use Sentinel-2~\cite{sentinel2} images for their superior spatiotemporal resolution and widespread use in vegetation monitoring~\cite{sentinel2, lang2023high, sialelli2024agbd, karaman2025gsr4b}. 
In addition to the $12$ spectral bands of Sentinel-2, \DATA carries an NDVI band, which is widely used as a proxy of vegetation productivity and biomass~\cite{ndvi}.
While the present work focuses on demonstrating the impact of dataset sampling on Sentinel-2, our dataset construction and sampling analysis could naturally be extended to other modalities relevant to ecology such as optical~\cite{wulder2016landsat, justice1998moderate}, SAR~\cite{torres2012gmes, rosenqvist2007alos}, LiDAR~\cite{dubayah2020gedi} sensors, or species~\cite{edwards2004gbif}, and climate~\cite{hersbach2020era5, chelsa} observations, which we leave for future work.

\vspace{-1em}
\paragraph{Patching.} 
Similar to previous works~\cite{seco, ssl4eo} we choose a patch size is $256 \times 256$ pixels ($2.56 \times 2.56$ km).
The exact amount of retrieved locations is $\num{254403}$,
each having up to $4$ dates, yielding a total of $1$M patches, for a final dataset size of $1.3$ TB.


\subsection{\SECO Model}
\label{sec:method_model}

In this section, we introduce our \SECO model, trained with a seasonal contrastive objective on \DATAx. 
We stress that our current focus is on pretraining dataset construction, not novel SSL training or backbone.

Our self-supervised training objective needs to capture spatial and seasonal patterns in our multi-temporal data. 
SSL4EO~\cite{ssl4eo} proposes a seasonal contrastive objective, but encourages learning season-agnostic features.
Instead, we use Seasonal Contrast (SeCo)~\cite{seco}, which learns both season-agnostic and season-specific representations, more appropriate for seasonality-sensitive tasks.
We use ResNet50 as our image encoding backbone, which has proven to be robust across a variety of remote sensing tasks~\cite{seco, satclip, wang2022self, ssl4eo}.
We dub \SECO our resulting model trained on \DATAx.
We also explore in \secref{sec:experiments} another version, \MOCOx, pretrained using the seasonal contrastive objective from SSL4EO.

We pretrain for $100$ epochs, with a batch size of $256$ on a single A100 GPU ($7$ days for \SECOx, $4$ days for \MOCOx).
See \secref{sec:details_model_implementation} for more implementation details.

\begin{figure}[t]
    \centering
    \includegraphics[width=\linewidth]{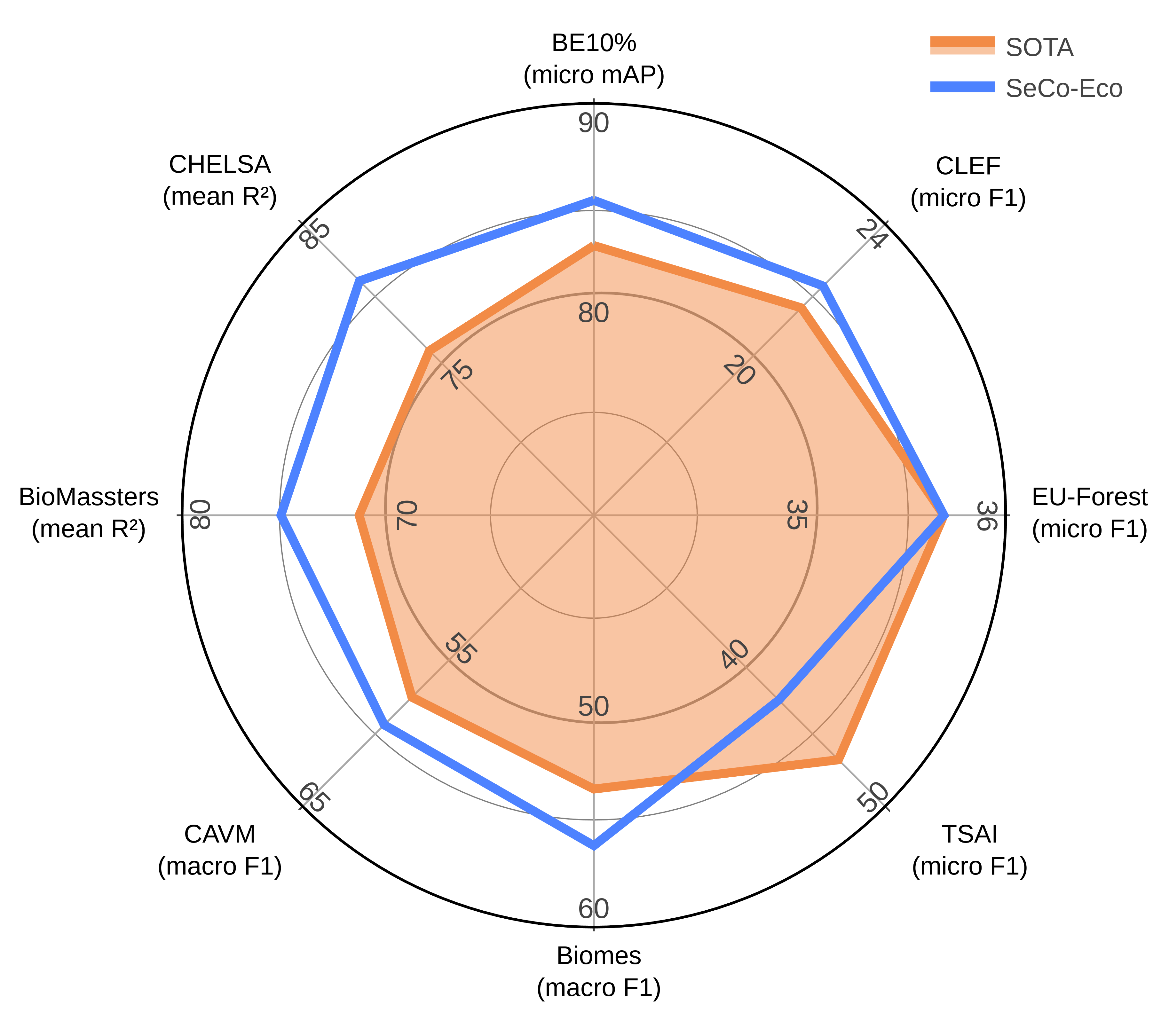}
    \caption{Linear Probing performance across all datasets. We compare \SECO against the respective best-performing model among our reported set of baselines.}
    \label{fig:spiderplot}
\end{figure}

\definecolor{lightblue}{RGB}{227, 234, 250}
\begin{table}[b]
    \centering
    \small
    \setlength{\tabcolsep}{4pt} 
    \renewcommand{\arraystretch}{1.2} 
    \begin{NiceTabular}{@{}lcccccccc@{}}\\
        \toprule 
        \multirow{3}{*}{Model} &
        \multicolumn{4}{c}{\makecell{Biomes\\(macro F1) }$ \uparrow$} &
        \multicolumn{4}{c}{\makecell{CAVM\\(macro F1) }$ \uparrow$}
        \\
        \cmidrule(lr){2-5} \cmidrule(lr){6-9} &
        \multicolumn{2}{c}{LP} &
        \multicolumn{2}{c}{10-NN} &
        \multicolumn{2}{c}{LP} &
        \multicolumn{2}{c}{20-NN}
        \\
        \midrule
        SeCo~\cite{seco} &
        \multicolumn{2}{c}{{\footnotesize $41.5 \pm 0.5$}} &
        \multicolumn{2}{c}{{\footnotesize $36.9 \pm 1.0$}} &
        \multicolumn{2}{c}{{\footnotesize $54.4 \pm 0.7$}} &
        \multicolumn{2}{c}{{\footnotesize $52.1 \pm 0.7$}}
        \\
        SatMAE~\cite{satmae} &
        \multicolumn{2}{c}{{\footnotesize $51.3 \pm 1.1$}} &
        \multicolumn{2}{c}{{\footnotesize $47.7 \pm 0.7$}} &
        \multicolumn{2}{c}{{\footnotesize $56.3 \pm 1.4$}} &
        \multicolumn{2}{c}{{\footnotesize $55.8 \pm 0.7$}}
        \\
        Satlas~\cite{saltas} &
        \multicolumn{2}{c}{{\footnotesize $48.3 \pm 1.6$}} &
        \multicolumn{2}{c}{{\footnotesize $47.6 \pm 0.9$}} &
        \multicolumn{2}{c}{{\footnotesize $53.8 \pm 2.0$}} &
        \multicolumn{2}{c}{{\footnotesize $53.2 \pm 0.5$}}
        \\
        Croma~\cite{croma} &
        \multicolumn{2}{c}{{\footnotesize $47.1 \pm 1.4$}} &
        \multicolumn{2}{c}{{\footnotesize $42.2 \pm 0.6$}} &
        \multicolumn{2}{c}{{\footnotesize $53.6 \pm 1.2$}} &
        \multicolumn{2}{c}{{\footnotesize $51.6 \pm 0.8$}}
        \\
        SSL4EO~\cite{ssl4eo} &
        \multicolumn{2}{c}{{\footnotesize $\underline{53.3} \pm 1.0$}} &
        \multicolumn{2}{c}{{\footnotesize $\underline{49.7} \pm 0.5$}} &
        \multicolumn{2}{c}{{\footnotesize $\underline{57.5} \pm 0.6$}} &
        \multicolumn{2}{c}{{\footnotesize $\underline{56.9} \pm 0.6$}}
        \\
        DOFA~\cite{dofa} &
        \multicolumn{2}{c}{{\footnotesize $49.7 \pm 1.3$}} &
        \multicolumn{2}{c}{{\footnotesize $42.9 \pm 0.5$}} &
        \multicolumn{2}{c}{{\footnotesize $56.4 \pm 1.6$}} &
        \multicolumn{2}{c}{{\footnotesize $53.5 \pm 0.6$}}
        \\
        \arrayrulecolor{black} \midrule
        \rowcolor{lightblue} \textbf{SeCo-Eco (ours)} &
        \multicolumn{2}{c}{{\footnotesize $\textbf{56.1} \pm 0.7$}} &
        \multicolumn{2}{c}{{\footnotesize $\textbf{51.1} \pm 0.9$}} &
        \multicolumn{2}{c}{{\footnotesize ${\textbf{59.4}} \pm 1.0$}} &
        \multicolumn{2}{c}{{\footnotesize $\textbf{59.5} \pm 0.8$}}
        \\
        \bottomrule
    \end{NiceTabular}
    \caption{
        Linear probing and K-Nearest Neighbor comparison of state of the art models with our \SECO pretrained on our \DATA on classification of two land cover datasets: global biomes and Arctic vegetation types \cite{cavm}.
        \textbf{Best}, \underline{second best}.
    }
    \label{tab:results_classification} 
\end{table}
\definecolor{lightblue}{RGB}{227, 234, 250}
\begin{table*}[t]
    \centering
    \small
    \setlength{\tabcolsep}{4pt} 
    \renewcommand{\arraystretch}{1.2} 
    \begin{NiceTabular}{@{}lcccccccccccccccc@{}}\\
        \toprule 
        \multirow{3}{*}{Model} &
        \multicolumn{4}{c}{\makecell{BE10\%\\(micro mAP) }$\uparrow$} &
        \multicolumn{4}{c}{\makecell{CLEF\\(micro F1) }$\uparrow$} &
        \multicolumn{4}{c}{\makecell{EU-Forest\\(micro F1) }$\uparrow$} &
        \multicolumn{4}{c}{\makecell{TSAI\\(micro F1) }$\uparrow$}
        \\
        \cmidrule(lr){2-5} \cmidrule(lr){6-9} \cmidrule(lr){10-13} \cmidrule(lr){14-17} &
        \multicolumn{2}{c}{LP} &
        \multicolumn{2}{c}{30-NN} &
        \multicolumn{2}{c}{LP} &
        \multicolumn{2}{c}{1-NN} &
        \multicolumn{2}{c}{LP} &
        \multicolumn{2}{c}{5-NN} &
        \multicolumn{2}{c}{LP} &
        \multicolumn{2}{c}{5-NN}
        \\
        \midrule
        SeCo~\cite{seco} &
        \multicolumn{2}{c}{{\footnotesize $79.2 \pm 0.0$}} &
        \multicolumn{2}{c}{{\footnotesize $77.8 \pm 0.1$}} &
        \multicolumn{2}{c}{{\footnotesize $20.8$}} &
        \multicolumn{2}{c}{{\footnotesize $12.3$}} &
        \multicolumn{2}{c}{{\footnotesize $31.3 \pm 0.7$}} &
        \multicolumn{2}{c}{{\footnotesize $30.6 \pm 0.2$}} &
        \multicolumn{2}{c}{{\footnotesize $23.4 \pm 0.0$}} &
        \multicolumn{2}{c}{{\footnotesize $35.2$}}
        \\
        SatMAE~\cite{satmae} &
        \multicolumn{2}{c}{{\footnotesize $79.7 \pm 0.2$}} &
        \multicolumn{2}{c}{{\footnotesize $79.6 \pm 0.0$}} &
        \multicolumn{2}{c}{{\footnotesize $21.6$}} &
        \multicolumn{2}{c}{{\footnotesize $\textbf{13.6}$}} &
        \multicolumn{2}{c}{{\footnotesize $\textbf{35.7} \pm 1.0$}} &
        \multicolumn{2}{c}{{\footnotesize $\textbf{33.3} \pm 0.1$}} &
        \multicolumn{2}{c}{{\footnotesize ${\textbf{46.8}} \pm 0.3$}} &
        \multicolumn{2}{c}{{\footnotesize{$\textbf{43.7}$}}}
        \\
        Satlas~\cite{saltas} &
        \multicolumn{2}{c}{{\footnotesize $77.9 \pm 0.2$}} &
        \multicolumn{2}{c}{{\footnotesize $77.9 \pm 0.0$}} &
        \multicolumn{2}{c}{{\footnotesize $18.9$}} &
        \multicolumn{2}{c}{{\footnotesize $11.8$}} &
        \multicolumn{2}{c}{{\footnotesize $30.0 \pm 0.2$}} &
        \multicolumn{2}{c}{{\footnotesize $30.0 \pm 0.2$}} &
        \multicolumn{2}{c}{{\footnotesize $42.9 \pm 0.0$}} &
        \multicolumn{2}{c}{{\footnotesize $40.8$}}
        \\
        Croma~\cite{croma} &
        \multicolumn{2}{c}{{\footnotesize $80.7 \pm 0.2$}} &
        \multicolumn{2}{c}{{\footnotesize $79.1 \pm 0.0$}} &
        \multicolumn{2}{c}{{\footnotesize $20.8$}} &
        \multicolumn{2}{c}{{\footnotesize $12.0$}} &
        \multicolumn{2}{c}{{\footnotesize $32.2 \pm 0.9$}} &
        \multicolumn{2}{c}{{\footnotesize $30.1 \pm 0.2$}} &
        \multicolumn{2}{c}{{\footnotesize $\underline{43.8} \pm 0.0$}} &
        \multicolumn{2}{c}{{\footnotesize $40.7$}}
        \\
        SSL4EO~\cite{ssl4eo} &
        \multicolumn{2}{c}{{\footnotesize $\underline{83.2} \pm 0.1$}} &
        \multicolumn{2}{c}{{\footnotesize $\underline{81.1} \pm 0.0$}} &
        \multicolumn{2}{c}{{\footnotesize $\underline{21.7}$}} &
        \multicolumn{2}{c}{{\footnotesize $12.6$}} &
        \multicolumn{2}{c}{{\footnotesize $32.6 \pm 0.1$}} &
        \multicolumn{2}{c}{{\footnotesize $31.5 \pm 0.2$}} &
        \multicolumn{2}{c}{{\footnotesize $42.3 \pm 0.0$}} &
        \multicolumn{2}{c}{{\footnotesize $\underline{40.9}$}}
        \\
        DOFA~\cite{dofa} &
        \multicolumn{2}{c}{{\footnotesize $80.1 \pm 0.0$}} &
        \multicolumn{2}{c}{{\footnotesize $77.3 \pm 0.1$}} &
        \multicolumn{2}{c}{{\footnotesize $20.3$}} &
        \multicolumn{2}{c}{{\footnotesize $12.1$}} &
        \multicolumn{2}{c}{{\footnotesize $\underline{34.8} \pm 0.9$}} &
        \multicolumn{2}{c}{{\footnotesize $30.0 \pm 0.3$}} &
        \multicolumn{2}{c}{{\footnotesize $35.1 \pm 0.0$}} &
        \multicolumn{2}{c}{{\footnotesize $37.4$}}
        \\
        \arrayrulecolor{black} \midrule
        \rowcolor{lightblue} \textbf{SeCo-Eco (ours)} &
        \multicolumn{2}{c}{{\footnotesize ${\textbf{85.3}} \pm 0.0$}} &
        \multicolumn{2}{c}{{\footnotesize ${\textbf{84.0}} \pm 0.0$}} &
        \multicolumn{2}{c}{{\footnotesize $\textbf{22.3}$}} &
        \multicolumn{2}{c}{{\footnotesize $\underline{13.0}$}} &
        \multicolumn{2}{c}{{\footnotesize $\textbf{35.7} \pm 0.4$}} &
        \multicolumn{2}{c}{{\footnotesize $\underline{32.4} \pm 0.2$}} &
        \multicolumn{2}{c}{{\footnotesize $42.7 \pm 0.0$}} &
        \multicolumn{2}{c}{{\footnotesize $40.6$}}
        \\
        \bottomrule
        \end{NiceTabular}
    \caption{Linear probing and K-Nearest Neighbor comparison of state of the art models with our \SECO pretrained on our \DATA on multi-label classification tasks. 
    CLEF and TSAI have official train and test splits, the standard deviation is only reported when relevant.
    \textbf{Best}, \underline{second best}.} 
    \label{tab:results_multitarget} 
\end{table*}

\section{Experiments}
\label{sec:experiments}

We present in \secref{sec:experiments_downstream_tasks} the downstream tasks we use for comparing geospatial foundation models, present experimental results in \secref{sec:experiments_results}, and ablations in \secref{sec:experiments_ablations}.

\subsection{Downstream Tasks and Evaluation} 
\label{sec:experiments_downstream_tasks}

Several benchmarks for geospatial foundation models have been proposed~\cite{yeh2021sustainbenchbenchmarks, lacoste2024geobenchtoward, fibaek2024phileobench, marsocci2024pangaeaa}, but none fit to our current setting of Sentinel-2, image-level representations for ecological applications.
Hence, we leverage existing datasets and propose new ones to evaluate the \DATA pretraining.

\vspace{-1em}
\paragraph{Protocol.}
We compare embeddings from \SECO with other geospatial foundation models operating on Sentinel-2 input data. 
Our choice of benchmarked methods is driven by the availability of reproducible code at the time of writing. 
For each model, we use the official implementation and adjust Sentinel-2 bands selection and normalization based on their respective pretraining setting.
Following Wang~\etal~\cite{ssl4eo}, we also crop or stretch the input patch size to align with the pretraining conditions.
We evaluate embeddings both with linear probing (LP) and K-Nearest Neighbor (k-NN) approaches.
For LP, we freeze the model backbone and train a single linear layer on top.
We use the AdamW~\cite{loshchilov2017decoupled} optimizer, train for up to a $\num{1000}$ epochs with early stopping, a learning rate of $1e^{-3}$, a batch size of $256$, on an NVIDIA T4 GPUs.
For k-NN, we follow existing literature~\cite{wu2018unsupervised, caron2021emerging} and aggregate labels from the k-nearest neighbors in the training set, based on cosine similarity. 
We use a softmax temperature of $0.07$ and grid-search $k$ for each task.
Unless specified otherwise for the task, we always split our data in $10$ test folds and randomly sample training and validation sets from the remaining data with $[0.9, 0.1]$ split. 
We report the mean and standard deviation for each metric across the $10$ test folds. The reported metric is picked based on common choices in the literature. See \secref{sec:detailed_results} for evaluations on a larger range of metrics and per-class results if applicable.


\vspace{-1em}
\paragraph{Classification Tasks.}
~\\
\noindent\textbf{\textit{Biomes.}}
We adapt the biomes task of Klemmer~\etal~\cite{satclip}, assembling a dataset of $52$k randomly selected inland locations and label them from a set of 15 classes according to Olson~\etal's biome map~\cite{biomes}. We adjust for latitude-longitude bias in the location selection. For each datapoint we download a $256\times256$ pixel ($2.56$ km) image from the least-clouded Sentinel-2 Harmonized dataset tile~\cite{sentinel2}. We choose images within a one month range from 15th of July/15th of January for the Northern/Southern hemisphere accordingly. We train using the cross entropy loss and report the macro F1 score. 

\noindent\textbf{\textit{Arctic Vegetation Types (CAVM).}}
We create an Arctic vegetation types task, as the Arctic ecosystem tends to be critically undersampled (see \tabrefshort{tab:related_work_datasets}). We assemble a dataset using $79$k randomly selected locations in equal area projection in the Arctic and label them according to the Arctic vegetation types CAVM dataset~\cite{cavm}. We use the broad map units (B, G, P, S, and W) as labels, resulting in $5$ vegetation categories. We choose images within a one month range from 15th of July. The downloaded satellite imagery, training, and metrics follow the setup of the biomes task.

\vspace{-1em}
\paragraph{Multi-Label Classification Tasks.}
~\\
\noindent\textbf{\textit{BigEarthNet.}}
BigEarthNet~\cite{bigearthnet} dataset is a $19$-class, multi-label land cover classification dataset.
It includes $590$k $1.2 \times 1.2$ km Sentinel-2 patches collected in 2017-2018 across Europe.
Although BigEarthNet is not specifically targeted for ecology, it is widely used for benchmarking GFMs, and we use it as a sanity check for the generalization power of our embeddings.
Following previous work, we report results on a predefined test set and use only 10\% of the remaining images for training~\cite{seco, ssl4eo, mmearth}.
We adapt the SSL4EO protocol~\cite{ssl4eo} for data preparation, train using a multi-label soft margin loss and measure performance by micro mean average precision.  

\noindent\textbf{\textit{GeoLifeCLEF 2023.}}
The GeoLifeCLEF 2023~\cite{geolifeclef} dataset contains $\num{5138}$ presence-absence surveys of $\num{2174}$ plant species across France and the United Kingdom.
Each survey reports all plant species found in a small plot (between $10$m$^2$ and $400$m$^2$).
For each location, we download a $1\times1$ km Sentinel-2 patch ($100\times100$ pixels).
We train with the binary cross-entropy loss, up weighting all presences by a factor of $12$ due to high imbalance between presences and absences.
We use the entire labeled dataset for training and communicate results on the official held-out test set.
We submit the predictions on the $22$k test surveys to the leaderboard and report the micro F1 score.

\noindent\textbf{\textit{{EU-Forest.}}}
We adapt the European 1 km-resolution tree occurrence dataset EU-Forest~\cite{euforest}
to a multi-label classification task. 
We sample $\num{51802}$ locations from the original data, covering $64$ species with at least $200$ occurrences, with some locations containing multiple species.
For each location, we download a $1\times1$ km Sentinel-2 patch. 
We train using a multi-label soft margin loss and measure performance by micro F1 score.  

\noindent\textbf{\textit{TreeSatAI.}}
The TreeSatAI~\cite{ahlswede2022treesatai} is a multimodal dataset for tree species identification with multi-label annotations for $15$ tree genera classes taken in Lower Saxony, Germany.
The dataset comprises $\num{50381}$ tiles of $60$ m width for several remote sensing products.
In our setting, we only use the Sentinel-2 $6\times6$ patches.
Similar to EU-Forest, we train with a multilabel soft margin loss and report the micro F1 score.
We communicate performance on the official test, and randomly select training and validation splits from the remaining data.

\vspace{-1em}
\paragraph{Regression Tasks.}
~\\    
\noindent\textbf{\textit{BioMassters.}}
BioMassters~\cite{nascetti2023biomassters} is a benchmark for aboveground biomass estimation in Finland from Sentinel-1/2 time series.
Initially designed for a dense pixel regression task, we reformulate it here as an image-level distribution prediction.
To this end, we divide the total distribution of biomass throughout the dataset into decile bins.
Since the first three bins account for zero biomass (\ie ground pixels), we merge them.
Then, for each $256\times256$ Sentinel-2  patch in the dataset, we compute the proportion of pixels falling into each of our $8$ bins.
Our model is tasked to predict the exact distribution of biomass for each image. 
Since the BioMassters dataset provides monthly images throughout the year, we split the task into a "summer" (June, July and August) and a "winter" (December, January and February) version, based on the season of the Sentinel-2 patches used as input.
We train using the Kullback-Leibler divergence and report the average coefficient of determination R$^2$ across bins as our main metric. 

\noindent\textbf{\textit{CHELSA Climate Regression.}}
Similar to SatClip~\cite{satclip} we propose to regress these aggregated climatic variables from pretrained geolocated embeddings.
CHELSA~\cite{chelsa} is a 1 km resolution global downscaled climate dataset, from which we extract the mean temperature (temp), total annual precipitation (prec), potential evaporation (evap) and site water balance (swb) from the 1981-2010 climatology of CHELSA v2.1~\cite{chelsa} for $50$k locations across the landmass.
For simplicity, we use the same locations and Sentinel-2 images as for the Biomes task.
After Gaussian-normalizing the values, we train using a mean squared error loss and use R$^2$ to measure performance.
    



\subsection{Results and Analysis}
\label{sec:experiments_results}

We compare the representation learned by \SECO on our \DATA across the above-defined tasks.
\figref{fig:spiderplot} summarizes the performance of \SECO in comparison to the strongest baseline on each task. 
Overall, we observe that \SECO outperforms all other approaches on all but one task, showing that a simple change in the sampling design of the pretraining dataset can yield significant improvements. 

\vspace{-1.2em}
\paragraph{Classification.}
\tabref{tab:results_classification}
\SECO outperforms all other methods on our classification tasks, both for linear probing and k-NN evaluation, followed by SSL4EO with $+2.8$ and $+1.9$ macro F1 LP performance gaps on the biomes and CAVM tasks, respectively.
The improvement of \SECO over SSL4EO can be explained by their difference in seasonal-contrastive training, as well as our dataset design (see \secref{sec:experiments_ablations} and \secref{sec:ablation_calendar} for more details).
The low performance of the RGB-based SeCo on biomes highlights the importance of multispectral images for the biomes classification task. 
The superior performance of \SECO over SSL4EO on CAVM illustrates the importance of including arctic regions in the pretraining set for ecological applications.

\vspace{-1em}
\paragraph{Multi-Label Classification.}
We compare performance on four multi-label classification tasks in \tabref{tab:results_multitarget}, three of which are specifically directed at predicting plant species communities. 
\SECO outperforms all other baselines in LP for BigEarthNet-10\% ($+2.1$ mAP), GeoLifeCLEF ($+0.6$ micro F1).
Interestingly, the largest performance gain from our approach is observed on the challenging BigEarthNet benchmark, which oversamples non-natural landscapes.
This indicates that despite its focus on capturing global phenological seasonality, the spatiotemporal distribution of \DATA still allows learning anthropic patterns.
On the other hand, \SECO performs $-4.1$ micro F1 below SatMAE on the TreeSatAI task, which we attribute to the small $6\times6$ patch size used for this task, which is far from the $224\times224$ both \SECO and SeCo are pretrained on. 

\vspace{-3em}
\paragraph{Regression.}
\definecolor{lightblue}{RGB}{227, 234, 250}
\begin{table}[h]
    \centering
    \small
    \setlength{\tabcolsep}{4pt} 
    \renewcommand{\arraystretch}{1.2} 
    \begin{NiceTabular}{@{}lcccccccc@{}}\\
        \toprule 
        \multirow{3}{*}{Model} &
        \multicolumn{4}{c}{\makecell{BioMassters\\(mean R\textsuperscript{2}) }$\uparrow$} &
        \multicolumn{4}{c}{\makecell{CHELSA\\(mean R\textsuperscript{2}) }$\uparrow$} \\
        \cmidrule(lr){2-5} \cmidrule(lr){6-9} 			& \multicolumn{2}{c}{LP} & \multicolumn{2}{c}{1-NN}
        & \multicolumn{2}{c}{LP} & \multicolumn{2}{c}{10-NN}\\
        \midrule
        SeCo~\cite{seco}& \multicolumn{2}{c}{{\footnotesize $51.2 \pm 0.0$}}& \multicolumn{2}{c}{{\footnotesize$-19.2$}}& \multicolumn{2}{c}{{\footnotesize $68.3 \pm 0.7$}}& \multicolumn{2}{c}{{\footnotesize$67.4 \pm 0.7$}}\\
        SatMAE~\cite{satmae}& \multicolumn{2}{c}{{\footnotesize $59.4 \pm 0.5$}}& \multicolumn{2}{c}{{\footnotesize$-18.0$}}& \multicolumn{2}{c}{{\footnotesize $\underline{76.3} \pm 0.6$}}& \multicolumn{2}{c}{{\footnotesize$77.6 \pm 0.7$}}\\
        Satlas~\cite{saltas}& \multicolumn{2}{c}{{\footnotesize $62.4 \pm 0.9$}}& \multicolumn{2}{c}{{\footnotesize$-17.8$}}& \multicolumn{2}{c}{{\footnotesize $68.3 \pm 0.9$}}& \multicolumn{2}{c}{{\footnotesize$73.3 \pm 0.7$}}\\
        Croma~\cite{croma}& \multicolumn{2}{c}{{\footnotesize $58.4 \pm 0.2$}}& \multicolumn{2}{c}{{\footnotesize$-18.1$}}& \multicolumn{2}{c}{{\footnotesize $73.3 \pm 0.9$}}& \multicolumn{2}{c}{{\footnotesize$71.2 \pm 0.5$}}\\
        SSL4EO~\cite{ssl4eo}& \multicolumn{2}{c}{{\footnotesize $\underline{71.3} \pm 0.1$}}& \multicolumn{2}{c}{{\footnotesize$\underline{-16.8}$}}& \multicolumn{2}{c}{{\footnotesize $75.8 \pm 0.6$}}& \multicolumn{2}{c}{{\footnotesize$\underline{77.7} \pm 0.5$}}\\
        DOFA~\cite{dofa}& \multicolumn{2}{c}{{\footnotesize $63.0 \pm 0.4$}}& \multicolumn{2}{c}{{\footnotesize$-18.3$}}& \multicolumn{2}{c}{{\footnotesize $69.6 \pm 0.6$}}& \multicolumn{2}{c}{{\footnotesize$70.7 \pm 0.7$}}\\
\arrayrulecolor{black} 			\midrule
        \rowcolor{lightblue} 			\textbf{SeCo-Eco (ours)}& \multicolumn{2}{c}{{\footnotesize ${\textbf{75.3}} \pm 0.3$}}& \multicolumn{2}{c}{{\footnotesize${\textbf{-16.3}}$}}& \multicolumn{2}{c}{{\footnotesize $\textbf{81.1} \pm 0.4$}}& \multicolumn{2}{c}{{\footnotesize$\textbf{81.0} \pm 0.5$}}\\
        \bottomrule
        \end{NiceTabular}
    \caption{Linear probing and K-Nearest Neighbor comparison of state of the art models with our \SECO pretrained on our \DATA on regression tasks. For the BioMassters task the standard deviation can only be reported for linear probing due to the fixed train and test sets.
    \textbf{Best}, \underline{second best}.} 
    \label{tab:results_regression} 
\end{table}
For the two regression tasks of BioMassters and CHELSA, we report in \tabref{tab:results_regression} the mean R$^2$ performance, aggregated across the BioMassters bins and CHELSA rasters.
\SECO outperforms all other baselines by a significant margin on both BioMassters ($+4.0$ R$^2$ LP) and CHELSA ($+4.8$ R$^2$ LP).
The large performance gap with respect to SSL4EO suggests that our model benefits from the more uniform spatial sampling of its pretraining dataset. 
Indeed, the BioMassters dataset is located in Finland, which is poorly covered by the SSL4EO pretraining dataset (\figrefshort{fig:spatial_sampling}).
Similarly, the CHELSA task requires uniform performance across the globe, which does not align with the urban-focused SSL4EO pretraining.
The negative R$^2$ scores on BioMassters indicate that 1-NN yields lower performance than a simple average prediction, suggesting that that this NN evaluation is not adapted to this task, for which linear probing should be preferred.
In comparison, the CHELSA task regresses climatic conditions, which evolve more smoothly throughout the models feature spaces, allowing to retrieve good estimates from neighboring embeddings.

\subsection{Ablation Study.}
\label{sec:experiments_ablations}



\definecolor{lightblue}{RGB}{227, 234, 250}
\begin{table*}[t]
\centering
\small
\setlength{\tabcolsep}{4pt} 
\renewcommand{\arraystretch}{1.2} 
\begin{NiceTabular}{@{}lcccccccccccccccc@{}}\\
    \toprule 
    \multirow{1}{*}{Model} &
    \multicolumn{2}{c}{\makecell{BE10\%\\(micro mAP) }$\uparrow$} &
    \multicolumn{2}{c}{\makecell{CLEF\\(micro F1) }$\uparrow$} &
    \multicolumn{2}{c}{\makecell{EU-Forest\\(micro F1) }$\uparrow$} &
    \multicolumn{2}{c}{\makecell{TSAI\\(micro F1) }$\uparrow$} &
    \multicolumn{2}{c}{\makecell{Biomes\\(macro F1) }$\uparrow$} &
    \multicolumn{2}{c}{\makecell{CAVM\\(macro F1) }$\uparrow$} &
    \multicolumn{2}{c}{\makecell{BioMassters\\(mean R\textsuperscript{2}) }$\uparrow$} &
    \multicolumn{2}{c}{\makecell{CHELSA\\(mean R\textsuperscript{2}) }$\uparrow$} \\
    \midrule
    SSL4EO~\cite{ssl4eo}& \multicolumn{2}{c}{{\footnotesize $83.2 \pm 0.1$}}& \multicolumn{2}{c}{{\footnotesize \underline{$21.7$}}}& \multicolumn{2}{c}{{\footnotesize $32.6 \pm 0.1$}}& \multicolumn{2}{c}{{\footnotesize \underline{$42.3$} $\pm 0.0$}}& \multicolumn{2}{c}{{\footnotesize $53.3 \pm 1.1$}}& \multicolumn{2}{c}{{\footnotesize $57.5$ $\pm 0.6$}}& \multicolumn{2}{c}{{\footnotesize $71.4 \pm 0.0$}}& \multicolumn{2}{c}{{\footnotesize $75.9 \pm 0.6$}}\\
    MoCo-Eco& \multicolumn{2}{c}{{\footnotesize \underline{$84.0$} $\pm 0.1$}}& \multicolumn{2}{c}{{\footnotesize \underline{$21.7$}}}& \multicolumn{2}{c}{{\footnotesize $\underline{35.4}$ $\pm 0.2$}}& \multicolumn{2}{c}{{\footnotesize $41.3 \pm 0.0$}}& \multicolumn{2}{c}{{\footnotesize $\textbf{58.4} \pm 0.8$}}& \multicolumn{2}{c}{{\footnotesize $\underline{59.1} \pm 0.7$}}& \multicolumn{2}{c}{{\footnotesize \underline{$73.4$} $\pm 0.1$}}& \multicolumn{2}{c}{{\footnotesize $\textbf{81.5} \pm 0.4$}}\\
    \arrayrulecolor{black} \midrule
    \rowcolor{lightblue} \textbf{SeCo-Eco (ours)}& \multicolumn{2}{c}{{\footnotesize $\textbf{85.3} \pm 0.0$}}& \multicolumn{2}{c}{{\footnotesize $\textbf{22.3}$}}& \multicolumn{2}{c}{{\footnotesize $\textbf{35.7} \pm 0.4$}}& \multicolumn{2}{c}{{\footnotesize $\textbf{42.7} \pm 0.0$}}& \multicolumn{2}{c}{{\footnotesize \underline{$56.1$} $\pm 0.7$}}& \multicolumn{2}{c}{{\footnotesize $\textbf{59.4} \pm 1.0$}}& \multicolumn{2}{c}{{\footnotesize $\textbf{75.2} \pm 0.1$}}& \multicolumn{2}{c}{{\footnotesize \underline{$81.1$} $\pm 0.4$}}\\
    \bottomrule
    \end{NiceTabular}
    \caption{
        Linear probing comparison of \MOCO and \SECO pretrained on \DATAx.
        \SECO learns both season-invariant and season-sensitive representations, which yield overall better performance than the season-invariant \MOCOx.
        \textbf{Best}, \underline{second best}.
    } 
    \label{tab:ablation_moco} 
\end{table*}

    \definecolor{lightblue}{RGB}{227, 234, 250}
    \begin{table}[t]
        \centering
        \small
        \setlength{\tabcolsep}{4pt} 
        \renewcommand{\arraystretch}{1.2} 
		\begin{NiceTabular}{@{}lcccccccc@{}}\\
            \toprule 
			\multirow{3}{*}{Model} &
			\multicolumn{4}{c}{\makecell{BioMassters S\\(mean R\textsuperscript{2}) }$\uparrow$} &
			\multicolumn{4}{c}{\makecell{BioMassters W\\(mean R\textsuperscript{2}) }$\uparrow$} \\
			\cmidrule(lr){2-5} \cmidrule(lr){6-9} 			& \multicolumn{2}{c}{LP} & \multicolumn{2}{c}{1-NN}
			& \multicolumn{2}{c}{LP} & \multicolumn{2}{c}{1-NN}\\
			\midrule
            SeCo~\cite{seco}& \multicolumn{2}{c}{{\footnotesize $51.3 \pm 0.0$}}& \multicolumn{2}{c}{{\footnotesize$-19.2$}}& \multicolumn{2}{c}{{\footnotesize $32.3 \pm 0.1$}}& \multicolumn{2}{c}{{\footnotesize$-30.6$}}\\
            SatMAE~\cite{satmae}& \multicolumn{2}{c}{{\footnotesize $59.5 \pm 0.6$}}& \multicolumn{2}{c}{{\footnotesize$-18.0$}}& \multicolumn{2}{c}{{\footnotesize $50.0 \pm 0.2$}}& \multicolumn{2}{c}{{\footnotesize$-26.3$}}\\
            Satlas~\cite{saltas}& \multicolumn{2}{c}{{\footnotesize $62.5 \pm 0.9$}}& \multicolumn{2}{c}{{\footnotesize$-17.8$}}& \multicolumn{2}{c}{{\footnotesize $51.7 \pm 1.1$}}& \multicolumn{2}{c}{{\footnotesize$-26.1$}}\\
            Croma~\cite{croma}& \multicolumn{2}{c}{{\footnotesize $58.5 \pm 0.2$}}& \multicolumn{2}{c}{{\footnotesize$-18.1$}}& \multicolumn{2}{c}{{\footnotesize $43.5 \pm 0.3$}}& \multicolumn{2}{c}{{\footnotesize$-27.0$}}\\
            SSL4EO~\cite{ssl4eo}& \multicolumn{2}{c}{{\footnotesize \underline{$71.4$} $\pm 0.0$}}& \multicolumn{2}{c}{{\footnotesize \underline{$-16.8$}}}& \multicolumn{2}{c}{{\footnotesize \underline{$63.2$} $\pm 0.1$}}& \multicolumn{2}{c}{{\footnotesize \underline{$-25.3$}}}\\
            DOFA~\cite{dofa}& \multicolumn{2}{c}{{\footnotesize $63.1 \pm 0.4$}}& \multicolumn{2}{c}{{\footnotesize$-18.3$}}& \multicolumn{2}{c}{{\footnotesize $55.0 \pm 0.4$}}& \multicolumn{2}{c}{{\footnotesize$-26.2$}}\\
\arrayrulecolor{black}          \midrule
            \rowcolor{lightblue}            \textbf{SeCo-Eco (ours)}& \multicolumn{2}{c}{{\footnotesize $\textbf{75.2} \pm 0.1$}}& \multicolumn{2}{c}{{\footnotesize$\textbf{-16.3}$}}& \multicolumn{2}{c}{{\footnotesize $\textbf{67.7} \pm 0.2$}}& \multicolumn{2}{c}{{\footnotesize$\textbf{-24.9}$}}\\
			\bottomrule
            \end{NiceTabular}
		\caption{
                Comparison of models using Summer (S) or Winter (W) images on BioMassters. Due to the fixed splits, the standard deviation can only be reported for linear probing.
                \textbf{Best}, \underline{second best}.
            } 
		\label{tab:ablation_winter} 
	\end{table}

\paragraph{Seasonal Pretraining.}
We compare in \tabref{tab:ablation_moco} the impact of pretraining on \DATA using the seasonal-contrastive objectives from SeCo~\cite{seco} and SSL4EO~\cite{ssl4eo} (\SECO and \MOCO models, respectively).
Our results show that \SECO features overall tend to perform on par or better than \MOCO features with linear probing, showing the benefit of learning not only season-agnostic features, but also season-specific ones.

\vspace{-1.em}
\paragraph{Winter Predictions.}
To test the influence of the acquisition date on model performance in downstream task, we compare the models on images taken from local winter months against summer months of BioMassters dataset. 
As shown in \tabref{tab:ablation_winter}, all models drop in performance when using the snow-covered winter images of BioMassters.
Still, we observe that \SECO clearly outperforms other models on both seasons, followed by SSL4EO, owing to their respective seasonal-contrastive pretraining.
Interestingly, the RGB-based SeCO model performs worst despite having the same pretraining strategy as \SECOx, suggesting multispectral imagery as critical to such tasks.
These results demonstrate the robustness of our learned phenologically-informed representation to seasonal changes.

\vspace{-1.em}
\paragraph{Limitations and Future Works.}
To recover less clouded images in each phenological season, we gather images across $2017$-$2024$, which may cause large temporal gaps between images of the same location, making our dataset inadequate for fine-grained temporal tasks.
Although not the focus of this work, pretraining more methods on \DATA besides \SECO and \MOCO would provide deeper insights into the respective merits of each.
Extending our dataset with additional modalities would likely allow learning richer features~\cite{mmearth, omnisat, anysat}, which we make possible by releasing all necessary metadata.
Finally, our dataset and model could naturally be used in a multi-modal contrastive learning framework aligning Sentinel-2 seasonal representations with text~\cite{silva2024large, yuan2024chatearthnet}, environmental variables~\cite{wildnet, glcwin}, or geolocation~\cite{geoclip, satclip}.


\section{Conclusion}

In this study, we propose a simple approach for sampling global seasonality-aware remote sensing datasets, from which we derive \DATAx, a multi-temporal Sentinel-2 dataset for pretraining geospatial foundation models targeted for macroecological applications.
Compared to previous works, our dataset uniformly samples the landmass and local phenological cycles.
We demonstrate that our simple spatiotemporal dataset sampling consistently improves the quality of self-supervised representations on a variety of macroecological tasks, highlighting the importance of pretraining set design, which could naturally be extended to additional relevant modalities.


\clearpage
{
    \section*{Acknowledgements}
This work made use of infrastructure services provided by the Science IT team of the University of Zurich (\url{www.s3it.uzh.ch}). 
We thank Benjamin Deneu for his helpful suggestions.
We acknowledge the Swiss National Science Foundation (SNF) for Advanced Grant Nr. TMAG-3\_216072.
Embed2Scale is co-funded by the EU Horizon Europe program under Grant Agreement No. $101131841$. Additional funding for this project has been provided by the Swiss State Secretariat for Education, Research and Innovation (SERI) and UK Research and Innovation (UKRI)

    \small
    \bibliographystyle{ieeenat_fullname}
    \bibliography{main}
}

\clearpage
\setcounter{page}{1}
\maketitlesupplementary

\renewcommand\thefigure{A-\arabic{figure}}
\renewcommand\thesection{A-\arabic{section}}
\renewcommand\thetable{A-\arabic{table}}
\renewcommand\theequation{A-\arabic{equation}}
\setcounter{equation}{0}
\setcounter{section}{0}
\setcounter{figure}{0}
\setcounter{table}{0}

\section{\DATA Dataset Construction}
\label{sec:details_dataset_construction}

In this section, we provide more details on our dataset construction protocol.

\paragraph{Spatial Sampling.}
We use the same approach as Major-TOM~\cite{majortom} for sampling locations uniformly across the landmass.
Our locations correspond to the center of the grid cells.

\paragraph{Seasonal Sampling.}
As explained in \secref{sec:method_dataset} and \figref{fig:temporal_sampling}, we define $4$ seasons as intervals between Greenup, Maturity, Senescence, Dormancy, and next Greenup variables.
The definition of these EVI variables can be found in \secref{sec:evi_definition}.
For each variable, we calculate the median day in the available years.
The EVI product from the MCD12Q2 v6.1~\cite{modis} product has missing values in non-vegetated and some evergreen areas (\eg tropics), for which we expect low seasonal variation.
We populate these with a nearest-neighbor approach by searching across geographical space.


For each location and season, we preselect all Sentinel-2 tiles across the $6$ years of data available $2017$-$2024$.
The broad range of years was chosen to account for high cloud coverage in some areas (\eg tropics in wet seasons).
Following previous work~\cite{ssl4eo}, we remove the tiles with less than less than $20$\% cloud coverage.
Finally, we choose the date and tile with the lowest cloud coverage for the location-season at hand.
If fewer than four seasonal images are available for a location due to cloud filtering, we use the $2$ or $3$ images that are available with less than $20$\% cloud coverage. 
Locations with only one image are excluded, accounting for $3$\% of initially sampled locations, mostly in the tropics and Antarctica.
Hence, final patches may be clouded, but the construction process ensures that the overall dataset has less than $20$\% cloud coverage.

We stress that the scope of this work is to study impact of spatiotemporal sampling compared to existing widely-used $4$-date seasonal datasets such as SeCo~\cite{seco} and SSL4EO~\cite{ssl4eo}.
As such, we follow the standard preprocessing procedure of these datasets regarding cloud filtering and the number of seasonal dates per year fair comparison across the computer vision literature.
However, realistic Earth Observation applications would require methods capable of handling arbitrarily sampled, potentially clouded, time series of satellite observations. 
We leave this exploration of the required dataset and models for further work. 

\paragraph{Data Source.}
Several open-access satellite products support vegetation monitoring.
\begin{itemize}
  \item Landsat missions~\cite{wulder2016landsat} offer a long-term multispectral record at $30$ m resolution, with a $16$-day revisit cycle (reduced to $8$ days since 2013).
  \item MODIS~\cite{justice1998moderate, salomonson1989modis} provides more spectral bands and a $1$\textendash$2$ day revisit rate, though at a coarser $250$\textendash$1000$ m resolution.
  \item Since $2015$, Sentinel-2~\cite{sentinel2} has been delivering $10$ m global imagery with a $5$-day maximum revisit period, balancing high spatial and temporal resolution. The Sentinel-2 instrument captures spectral bands indicative of ecological patterns, such as red-edge wavelengths sensitive to vegetation stress and chlorophyll content~\cite{delegido2011meris}.
  \item Radar sensors may provide diverse ecological insights depending on their frequency: C-band such as Sentinel-1~\cite{torres2012gmes} detects foliage, topography, and moisture, while L-band such as ALOS PALSAR~\cite{rosenqvist2007alos} can characterize wood structure.
\end{itemize}

In this paper, we chose Sentinel-2 due to its widespread use for large-scale vegetation monitoring~\cite{lang2023high, sialelli2024agbd, karaman2025gsr4b}, but we believe our conclusions remain applicable and may be extended to other satellite products in future works.
We leave the exploration of our proposed spatiotemporal sampling for multimodal representation learning for future work. 

\paragraph{Downloading.}
The \DATA dataset is downloaded from Google Earth Engine using code from SeCo~\cite{seco} and SSL4EO-S12~\cite{ssl4eo} with altered data source, seasonality, and data distribution.
We use the Sentinel-2A MSI collection which, compared to Sentinel-2C, has atmospheric correction and depicts more accurately features on the ground~\cite{sentinel2}.
We use harmonized version of the product instead of the original one, as it corrects for normalization issues in 2022.
We use Sentinel tiles with less than 20\% cloud coverage.


\section{Implementation Details}
\label{sec:details_model_implementation}

In this section, we provide more details on the implementation and training of our models.

\paragraph{Input Bands.}
Our models \SECO and \MOCO are trained to take as input the $8$ Sentinel-2 bands for ecological applications.
Specifically, we use the B2, B3, B4, B5, B6, B7, B8, and B8A bands.
While B2-B4 provide information on foliage color, which helps to assess seasonality and plant health, B5-B7 capture red-edge wavelengths sensitive to vegetation stress and chlorophyll content, and B8 and B8A in near-infrared range are useful to distinguish non-vegetated areas. 
In addition, we also include the NDVI index as a remote sensing-based proxy of vegetation productivity and biomass~\cite{ndvi}.
As a result, our models expect 9 channels as input.

We leave the exploration of pretraining on our \DATA sampling with more bands or modalities for future work.

\paragraph{Weighted Sampling.}
Despite the uniform global sampling of \DATAx, some locations may have more interesting geographical and seasonal dynamics than others.
In order to drive the pretraining towards regions with richer ecological patterns, we use a weighted sampling in our pretraining dataloader.
Specifically, we assign a $\div4$ weight to non-vegetated areas, identified as mean NDVI $<0.1$ in all seasons ($17$\% of \DATAx), focusing less on deserts and ice packs.
We oversample mountain regions with a $\times2$ weight, identified with the GMBA Mountain Inventory~\cite{gmba} ($16$\% of \DATAx), focusing more on ecologically diverse areas, as mountain regions harbor the highest diversity and heterogeneity of ecoregions.


\paragraph{Pretraining.}
We pretrain \SECO using the hyperparameters and code provided by Ma{\~n}as~\etal~\cite{seco}, using MoCo v2~\cite{mocov2}, with minor changes: we replace the RGB input with multispectral images and set the length of the negative examples queue to $\num{65536}$, following the implementation of Wang~\etal~\cite{ssl4eo}.

We pretrain \MOCO using the hyperparameters and code provided by Wang~\etal~\cite{ssl4eo}, adapted for a single A100 GPU with batch size of $256$.

Finally, we modify the random seasonal sampling found in the implementations of SeCo~\cite{seco} and SSL4EO~\cite{ssl4eo}.
When randomly selecting seasons at batch construction time, both use:~\\
\texttt{np.random.choice(..., replace=True)}~,~\\
although we believe:~\\
\texttt{np.random.choice(..., replace=False)}~\\
is the correct implementation of their respective methods, as this avoids contrasting an image against itself.


\section{EVI-based Seasonality}
\label{sec:evi_definition}

\begin{figure}[t]
    \centering
    \includegraphics[width=\linewidth]{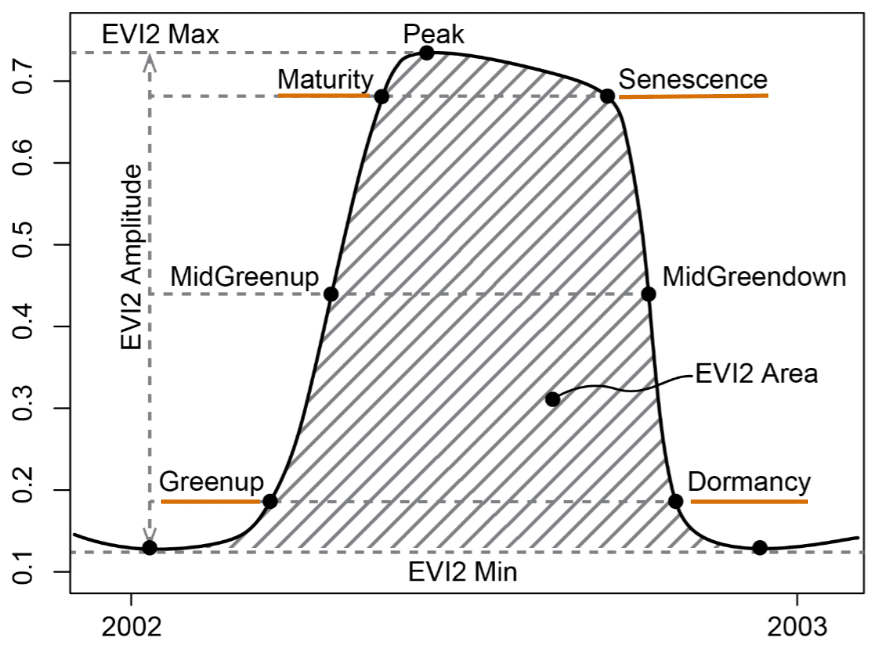}
    \caption{
    Enhanced Vegetation Index (EVI) curve of the vegetation cycle at a given location.
    Based on this curve, the Greenup, Maturity, Senescence, and Dormancy seasonality variables are defined as detailed in \tabrefshort{tab:evi_definition}.
    Image taken from~\cite{mcduserguide}.
    }
    \label{fig:evi_definition}
\end{figure}

\begin{table}
    \small
    \centering
    
    \begin{tabular}{@{}lc@{}}
    \toprule    
    {Name} & {Definition - Date when...} \\
    \midrule
    Greenup & EVI first crossed 15\% of segment EVI amplitude \\
    Maturity & EVI first crossed 90\% of segment EVI amplitude   \\
    Senescence & EVI last crossed 90\% of segment EVI amplitude  \\
    Dormancy & EVI last crossed 15\% of segment EVI amplitude  \\
    \bottomrule
    \end{tabular}
   
    \caption{
        Definition of the Greenup, Maturity, Senescence, and Dormancy seasonality variables based on the EVI curve (\figrefshort{fig:evi_definition}).
    }
    \label{tab:evi_definition}
\end{table}

We use the Enhanced Vegetation Index (EVI) from the MCD12Q2 v6.1~\cite{modis} product of the MODIS~\cite{justice1998moderate} satellite mission to define our local, phenology-informed seasons.
Similar to NDVI, the EVI index is commonly used to quantify the greenness of an area, but is more sensitive in areas with dense vegetation cover.
\figref{fig:evi_definition} illustrate a typical EVI curve over the year, and \tabref{tab:evi_definition} details how the Greenup, Maturity, Senescence, and Dormancy seasonality variables are defined.
For each location in our dataset, we choose 4 images, one for each season, close to the middle between the four EVI-derived variables.
See the MCD12Q2 user guide~\cite{mcduserguide} for more details on EVI variables.


\section{Calendar Ablation}
\label{sec:ablation_calendar}

\definecolor{lightblue}{RGB}{227, 234, 250}
\begin{table*}[t]
\centering
\small
\setlength{\tabcolsep}{4pt} 
\renewcommand{\arraystretch}{1.2} 
\begin{NiceTabular}{@{}lcccccccccccccccc@{}}\\
    \toprule 
    \multirow{1}{*}{Model} &
    \multicolumn{2}{c}{\makecell{BE10\%\\(micro mAP) }$\uparrow$} &
    \multicolumn{2}{c}{\makecell{CLEF\\(micro F1) }$\uparrow$} &
    \multicolumn{2}{c}{\makecell{EU-Forest\\(micro F1) }$\uparrow$} &
    \multicolumn{2}{c}{\makecell{TSAI\\(micro F1) }$\uparrow$} &
    \multicolumn{2}{c}{\makecell{Biomes\\(macro F1) }$\uparrow$} &
    \multicolumn{2}{c}{\makecell{CAVM\\(macro F1) }$\uparrow$} &
    \multicolumn{2}{c}{\makecell{BioMassters\\(mean R\textsuperscript{2}) }$\uparrow$} &
    \multicolumn{2}{c}{\makecell{Chelsa\\(mean R\textsuperscript{2}) }$\uparrow$} \\
    \midrule
    SeCo-Calendar &
    \multicolumn{2}{c}{{\footnotesize $\textbf{85.3} \pm 0.0$}} &
    \multicolumn{2}{c}{\footnotesize $22.4$} &
    \multicolumn{2}{c}{\footnotesize $34.2 \pm 0.1$} &
    \multicolumn{2}{c}{\footnotesize $40.8 \pm 0.0$} &
    \multicolumn{2}{c}{\footnotesize $55.2 \pm 1.0$} &
    \multicolumn{2}{c}{\footnotesize $58.7 \pm 0.8$} &
    \multicolumn{2}{c}{\footnotesize $\textbf{75.7} \pm 0.0$} &
    \multicolumn{2}{c}{\footnotesize $80.6 \pm 0.5$}
    \\
    \arrayrulecolor{black} \midrule
    \rowcolor{lightblue} \textbf{SeCo-Eco (ours)} &
    \multicolumn{2}{c}{{\footnotesize $\textbf{85.3} \pm 0.0$}} &
    \multicolumn{2}{c}{{\footnotesize $\textbf{22.7}$}} &
    \multicolumn{2}{c}{\footnotesize $\textbf{35.7} \pm 0.4$} &
    \multicolumn{2}{c}{\footnotesize $\textbf{42.7} \pm 0.0$} &
    \multicolumn{2}{c}{\footnotesize $\textbf{56.1} \pm 0.7$} &
    \multicolumn{2}{c}{\footnotesize $\textbf{59.4} \pm 1.0$} &
    \multicolumn{2}{c}{\footnotesize $75.1 \pm 0.0$} &
    \multicolumn{2}{c}{\footnotesize $\textbf{81.1} \pm 0.4$}
    \\
    \bottomrule
    \end{NiceTabular}
    \caption{
        Linear probing comparison of \SECO and SeCo-Calendar pretrained on EVI-based and calendar-based seasonal samplings, respectively.
        EVI-based samplings overally yields better features for downstream macroecological tasks, with the exception of the BioMassters dataset.
        \textbf{Best}.
    } 
    \label{tab:ablation_calendar} 
\end{table*}

Our temporal sampling of \DATA described in \secref{sec:method_dataset} makes the assumption that pretraining on EVI-based seasonal samplings rather than calendar seasons yields better features for ecological downstream tasks. 
To verify this claim, we assemble the \DATAx-Calendar dataset, which follows the same spatial sampling as \DATAx, but with a temporal sampling based on calendar dates following SSL4EO-S12~\cite{ssl4eo}.
We derive SeCo-Calendar from this dataset, by using the same pretraining recipe and backbone as for our \SECOx, and compare in \tabref{tab:ablation_calendar} their respective performance across downstream tasks.
We observe that our proposed EVI-based seasonal sampling yields representations which overall perform better than calendar-based sampling on most downstream tasks. 
In particular, EU-Forest ($+1.5$ micro F1), TSAI ($+1.9$ macro F1), and Biomes ($+0.9$ macro F1) prove to benefit from the finer phenology-informed features of \SECOx.
These results validate the importance of temporal sampling and the definition of local seasonality to capture local ecological patterns.


\section{Downstream Tasks}
\label{sec:downstream_tasks_sampling}

\begin{figure*}[t]  
    \centering
    \begin{subfigure}{0.49\textwidth}
        \includegraphics[width=\linewidth]{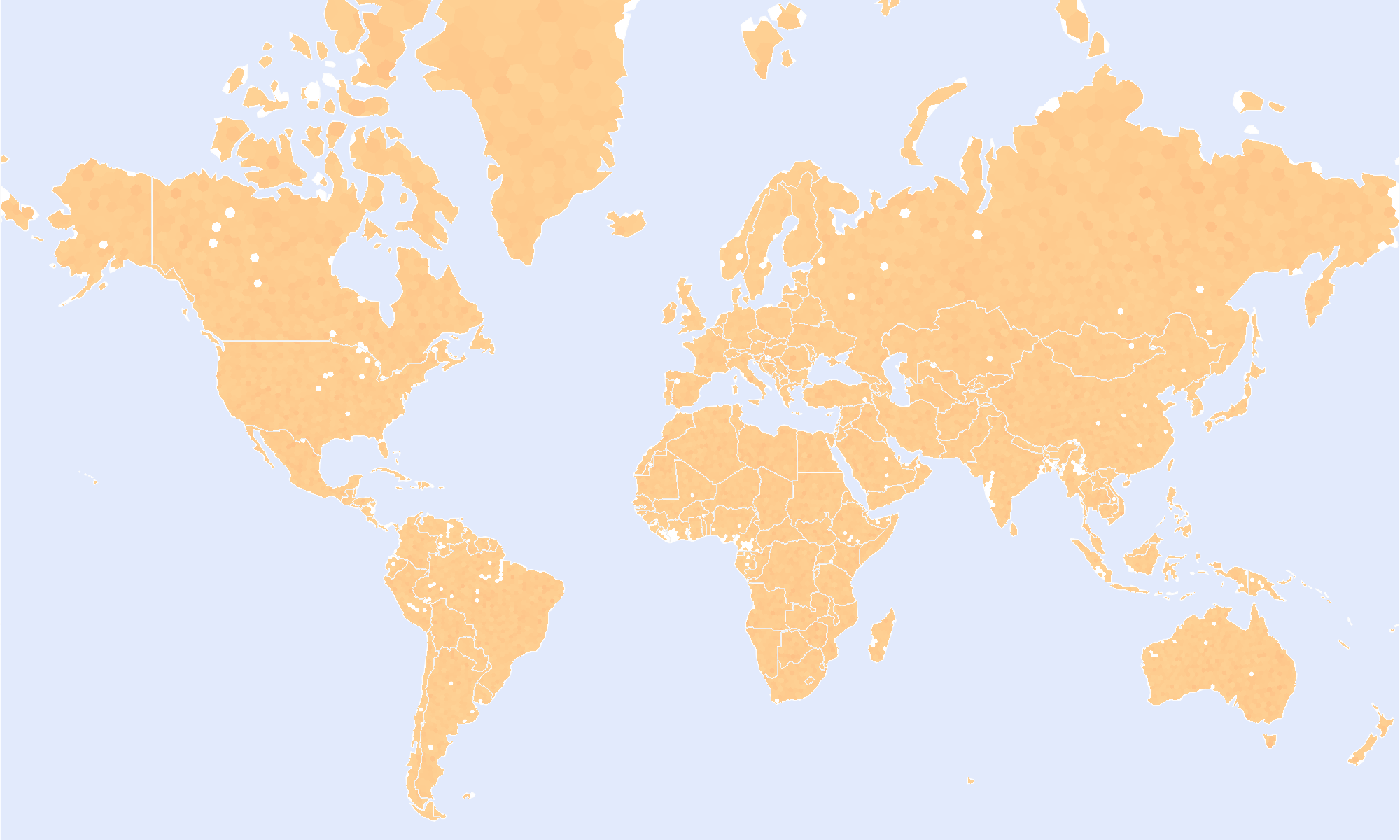}
        \caption{Biomes}
    \end{subfigure}
    \hfill
    \begin{subfigure}{0.49\textwidth}
        \includegraphics[width=\linewidth]{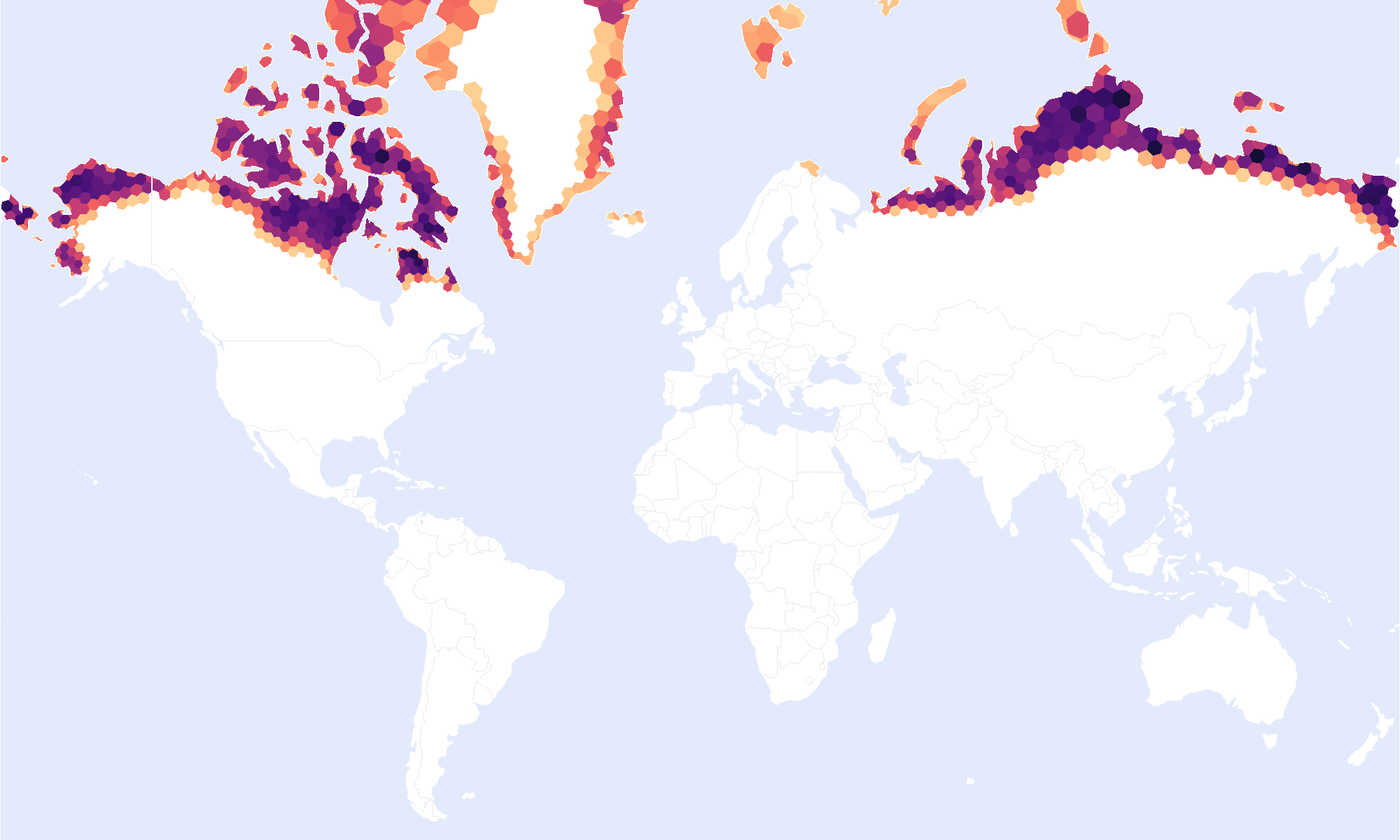}
        \caption{CAVM}
    \end{subfigure}

    \vspace{0.5em}  

    \begin{subfigure}{0.49\textwidth}
        \includegraphics[width=\linewidth]{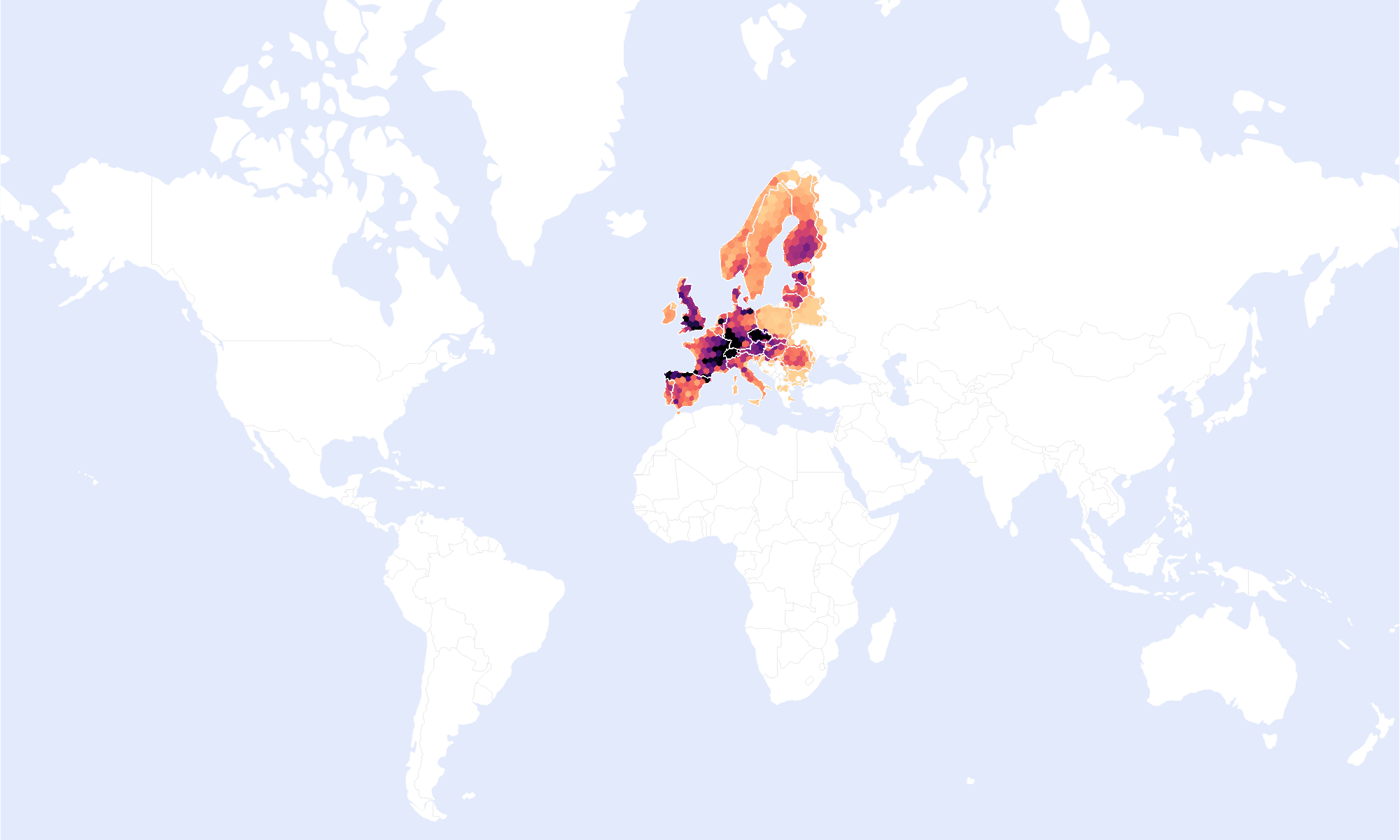}
        \caption{EU-Forest}
    \end{subfigure}
    \hfill
    \begin{subfigure}{0.49\textwidth}
        \includegraphics[width=\linewidth]{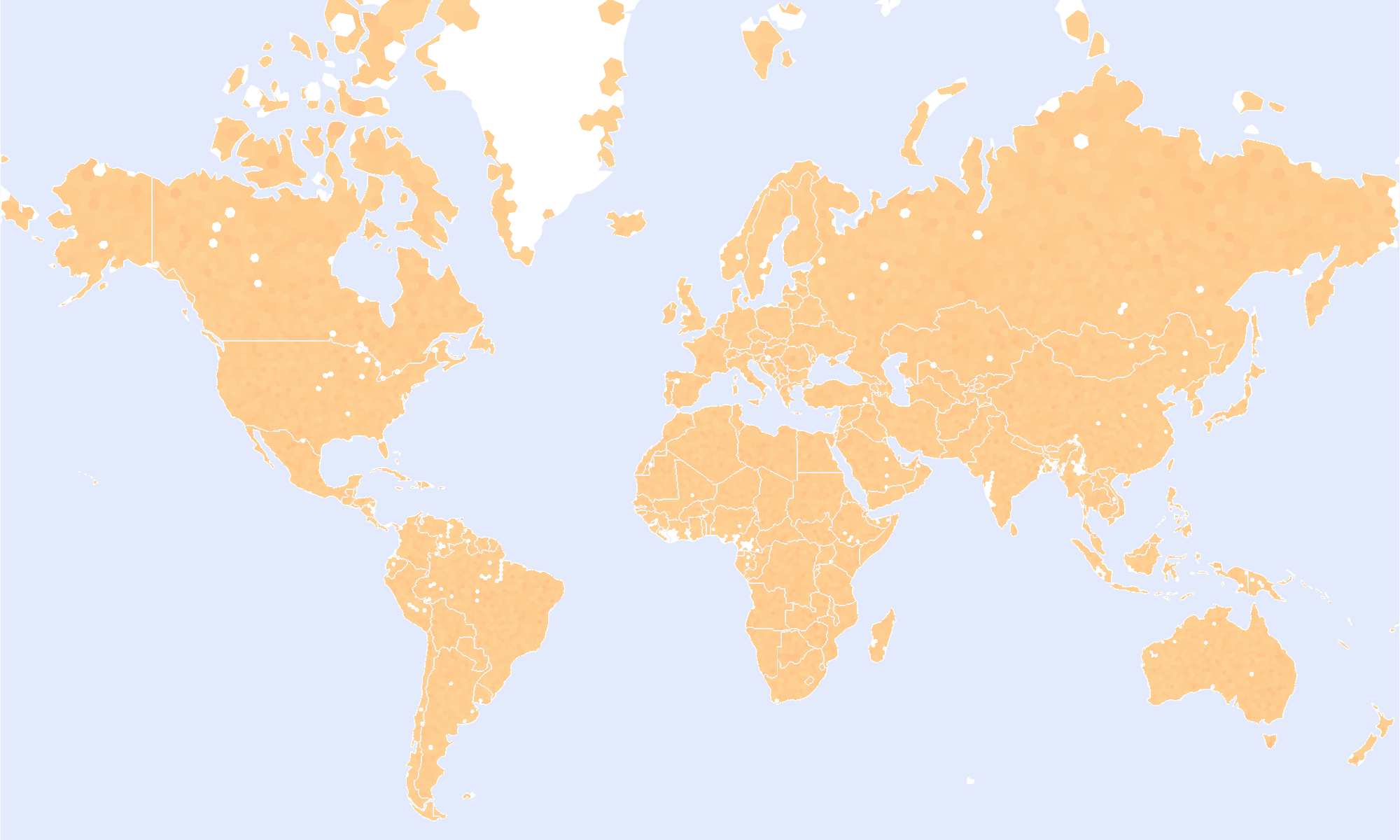}
        \caption{CHELSA}
    \end{subfigure}

    \caption{
        Spatial distribution of the four new downstream tasks created for this work. 
        We sample Biomes and CHELSA locations uniformly across the landmass.
        Meanwhile, the CAVM dataset is located in arctic regions and EU-Forest is limited to Europe.
    }
    \label{fig:spatial_sampling_downstream_tasks}
\end{figure*}

We illustrate in \figref{fig:spatial_sampling_downstream_tasks} the spatial distribution of the samplings used for the new downstream tasks proposed in this paper: Biomes, CAVM, EU-Forest, and CHELSA


\section{Detailed Results}
\label{sec:detailed_results}

Beyond evaluating performance with the most established metric per dataset, we provide further experimental results on an expanded set of metrics. 

\definecolor{lightblue}{RGB}{227, 234, 250}
\begin{table*}[ht]
    \centering
    \small
    \setlength{\tabcolsep}{4pt} 
    \renewcommand{\arraystretch}{1.2} 
    \begin{NiceTabular}{@{}lcccccccccccccccc@{}}\\
        \toprule 
        \multirow{5}{*}{Model} &
        \multicolumn{16}{c}{BE10\%~\cite{bigearthnet}}
        \\
        \cmidrule(lr){2-17} &
        \multicolumn{4}{c}{Macro F1 $\uparrow$} &
        \multicolumn{4}{c}{Micro F1 $\uparrow$} &
        \multicolumn{4}{c}{Macro mAP $\uparrow$} &
        \multicolumn{4}{c}{Micro mAP $\uparrow$}
        \\
        \cmidrule(lr){2-5} \cmidrule(lr){6-9} \cmidrule(lr){10-13} \cmidrule(lr){14-17} &
        \multicolumn{2}{c}{LP} &
        \multicolumn{2}{c}{30-NN} &
        \multicolumn{2}{c}{LP} &
        \multicolumn{2}{c}{30-NN} &
        \multicolumn{2}{c}{LP} &
        \multicolumn{2}{c}{30-NN} &
        \multicolumn{2}{c}{LP} &
        \multicolumn{2}{c}{30-NN}
        \\
        \midrule
        SeCo~\cite{seco} &
        \multicolumn{2}{c}{{\footnotesize $56.3 \pm 0.3$}} &
        \multicolumn{2}{c}{{\footnotesize $36.0 \pm 0.1$}} &
        \multicolumn{2}{c}{{\footnotesize $68.9 \pm 0.2$}} &
        \multicolumn{2}{c}{{\footnotesize $44.7 \pm 0.1$}} &
        \multicolumn{2}{c}{{\footnotesize $64.5 \pm 0.2$}} &
        \multicolumn{2}{c}{{\footnotesize $62.4 \pm 0.2$}} &
        \multicolumn{2}{c}{{\footnotesize $79.2 \pm 0.0$}} &
        \multicolumn{2}{c}{{\footnotesize $77.8 \pm 0.1$}}
        \\
        SatMAE~\cite{satmae} &
        \multicolumn{2}{c}{{\footnotesize $58.9 \pm 0.7$}} &
        \multicolumn{2}{c}{{\footnotesize $39.0 \pm 0.1$}} &
        \multicolumn{2}{c}{{\footnotesize $69.3 \pm 0.3$}} &
        \multicolumn{2}{c}{{\footnotesize $47.5 \pm 0.1$}} &
        \multicolumn{2}{c}{{\footnotesize $66.2 \pm 0.3$}} &
        \multicolumn{2}{c}{{\footnotesize $65.1 \pm 0.2$}} &
        \multicolumn{2}{c}{{\footnotesize $79.7 \pm 0.2$}} &
        \multicolumn{2}{c}{{\footnotesize $79.6 \pm 0.0$}}
        \\
        Satlas~\cite{saltas} &
        \multicolumn{2}{c}{{\footnotesize $55.7 \pm 1.2$}} &
        \multicolumn{2}{c}{{\footnotesize $37.3 \pm 0.1$}} &
        \multicolumn{2}{c}{{\footnotesize $67.3 \pm 0.7$}} &
        \multicolumn{2}{c}{{\footnotesize $45.9 \pm 0.1$}} &
        \multicolumn{2}{c}{{\footnotesize $64.8 \pm 0.2$}} &
        \multicolumn{2}{c}{{\footnotesize $62.2 \pm 0.2$}} &
        \multicolumn{2}{c}{{\footnotesize $77.9 \pm 0.2$}} &
        \multicolumn{2}{c}{{\footnotesize $77.9 \pm 0.0$}}
        \\
        Croma~\cite{croma} &
        \multicolumn{2}{c}{{\footnotesize $59.9 \pm 0.5$}} &
        \multicolumn{2}{c}{{\footnotesize $37.2 \pm 0.1$}} &
        \multicolumn{2}{c}{{\footnotesize $70.7 \pm 0.2$}} &
        \multicolumn{2}{c}{{\footnotesize $46.1 \pm 0.1$}} &
        \multicolumn{2}{c}{{\footnotesize $67.1 \pm 0.1$}} &
        \multicolumn{2}{c}{{\footnotesize $63.6 \pm 0.3$}} &
        \multicolumn{2}{c}{{\footnotesize $80.7 \pm 0.2$}} &
        \multicolumn{2}{c}{{\footnotesize $79.1 \pm 0.0$}}
        \\
        SSL4EO~\cite{ssl4eo} &
        \multicolumn{2}{c}{{\footnotesize $\underline{63.1} \pm 0.2$}} &
        \multicolumn{2}{c}{{\footnotesize $\underline{39.6} \pm 0.1$}} &
        \multicolumn{2}{c}{{\footnotesize $\underline{72.5} \pm 0.2$}} &
        \multicolumn{2}{c}{{\footnotesize $\underline{47.9} \pm 0.1$}} &
        \multicolumn{2}{c}{{\footnotesize $\underline{71.1} \pm 0.3$}} &
        \multicolumn{2}{c}{{\footnotesize $\underline{67.8} \pm 0.2$}} &
        \multicolumn{2}{c}{{\footnotesize $\underline{83.2} \pm 0.1$}} &
        \multicolumn{2}{c}{{\footnotesize $\underline{81.1} \pm 0.0$}}
        \\
        DOFA~\cite{dofa} &
        \multicolumn{2}{c}{{\footnotesize $59.9 \pm 0.6$}} &
        \multicolumn{2}{c}{{\footnotesize $37.8 \pm 0.2$}} &
        \multicolumn{2}{c}{{\footnotesize $70.1 \pm 0.2$}} &
        \multicolumn{2}{c}{{\footnotesize $46.1 \pm 0.1$}} &
        \multicolumn{2}{c}{{\footnotesize $66.9 \pm 0.2$}} &
        \multicolumn{2}{c}{{\footnotesize $62.7 \pm 0.2$}} &
        \multicolumn{2}{c}{{\footnotesize $80.1 \pm 0.0$}} &
        \multicolumn{2}{c}{{\footnotesize $77.3 \pm 0.1$}}
        \\
        \arrayrulecolor{black} \midrule
        \rowcolor{lightblue} \textbf{SeCo-Eco (ours)} &
        \multicolumn{2}{c}{{\footnotesize $\textbf{66.8} \pm 0.3$}} &
        \multicolumn{2}{c}{{\footnotesize $\textbf{41.4} \pm 0.1$}} &
        \multicolumn{2}{c}{{\footnotesize $\textbf{75.0} \pm 0.1$}} &
        \multicolumn{2}{c}{{\footnotesize $\textbf{49.9} \pm 0.1$}} &
        \multicolumn{2}{c}{{\footnotesize $\textbf{74.1} \pm 0.2$}} &
        \multicolumn{2}{c}{{\footnotesize $\textbf{71.7} \pm 0.2$}} &
        \multicolumn{2}{c}{{\footnotesize $\textbf{85.3} \pm 0.0$}} &
        \multicolumn{2}{c}{{\footnotesize $\textbf{84.0} \pm 0.0$}}
        \\
        \bottomrule
    \end{NiceTabular}
    \caption{
        Linear probing and K-Nearest Neighbor performance across multiple metrics for the BigEarthNet-10\% task.
        \textbf{Best}, \underline{second best}.
    } 
    \label{tab:all_metrics_bigearthnet} 
\end{table*}
\definecolor{lightblue}{RGB}{227, 234, 250}
\begin{table*}[h]
    \centering
    \small
    \setlength{\tabcolsep}{4pt} 
    \renewcommand{\arraystretch}{1.2} 
    \begin{NiceTabular}{@{}lcccccccccccccccc@{}}\\
        \toprule 
        \multirow{5}{*}{Model} &
        \multicolumn{16}{c}{EU-Forest~\cite{euforest}} 
        \\
        \cmidrule(lr){2-17} &
        \multicolumn{4}{c}{Macro AUROC $\uparrow$} &
        \multicolumn{4}{c}{Macro F1 $\uparrow$} &
        \multicolumn{4}{c}{Micro AUROC $\uparrow$} &
        \multicolumn{4}{c}{Micro F1 $\uparrow$} 
        \\
        \cmidrule(lr){2-5} \cmidrule(lr){6-9} \cmidrule(lr){10-13} \cmidrule(lr){14-17} & \multicolumn{2}{c}{LP} & \multicolumn{2}{c}{5-NN} &
        \multicolumn{2}{c}{LP} & \multicolumn{2}{c}{5-NN} &
        \multicolumn{2}{c}{LP} & \multicolumn{2}{c}{5-NN} &
        \multicolumn{2}{c}{LP} & \multicolumn{2}{c}{5-NN}
        \\
        \midrule
        SeCo~\cite{seco} &
        \multicolumn{2}{c}{{\footnotesize $82.6 \pm 0.0$}} &
        \multicolumn{2}{c}{{\footnotesize $63.9 \pm 0.3$}} &
        \multicolumn{2}{c}{{\footnotesize $12.3 \pm 0.7$}} &
        \multicolumn{2}{c}{{\footnotesize $18.2 \pm 0.3$}} &
        \multicolumn{2}{c}{{\footnotesize $90.6 \pm 0.1$}} &
        \multicolumn{2}{c}{{\footnotesize $77.6 \pm 0.2$}} &
        \multicolumn{2}{c}{{\footnotesize $31.3 \pm 0.9$}} &
        \multicolumn{2}{c}{{\footnotesize $30.6 \pm 0.2$}}
        \\
        SatMAE~\cite{satmae} &
        \multicolumn{2}{c}{{\footnotesize $\underline{84.6} \pm 0.2$}} &
        \multicolumn{2}{c}{{\footnotesize $\textbf{66.7} \pm 0.4$}} &
        \multicolumn{2}{c}{{\footnotesize $\textbf{15.0} \pm 0.7$}} &
        \multicolumn{2}{c}{{\footnotesize $\textbf{21.0} \pm 0.3$}} &
        \multicolumn{2}{c}{{\footnotesize $\underline{91.6} \pm 0.1$}} &
        \multicolumn{2}{c}{{\footnotesize $\textbf{79.8} \pm 0.2$}} &
        \multicolumn{2}{c}{{\footnotesize $\textbf{35.7} \pm 0.9$}} &
        \multicolumn{2}{c}{{\footnotesize $\textbf{33.3} \pm 0.1$}}
        \\
        Satlas~\cite{saltas} &
        \multicolumn{2}{c}{{\footnotesize $81.1 \pm 0.3$}} &
        \multicolumn{2}{c}{{\footnotesize $62.7 \pm 0.3$}} &
        \multicolumn{2}{c}{{\footnotesize $10.1 \pm 0.4$}} &
        \multicolumn{2}{c}{{\footnotesize $17.5 \pm 0.3$}} &
        \multicolumn{2}{c}{{\footnotesize $89.6 \pm 0.1$}} &
        \multicolumn{2}{c}{{\footnotesize $76.7 \pm 0.2$}} &
        \multicolumn{2}{c}{{\footnotesize $29.8 \pm 1.5$}} &
        \multicolumn{2}{c}{{\footnotesize $30.0 \pm 0.2$}}
        \\
        Croma~\cite{croma} &
        \multicolumn{2}{c}{{\footnotesize $82.9 \pm 0.3$}} &
        \multicolumn{2}{c}{{\footnotesize $63.6 \pm 0.3$}} &
        \multicolumn{2}{c}{{\footnotesize $12.2 \pm 0.7$}} &
        \multicolumn{2}{c}{{\footnotesize $18.1 \pm 0.3$}} &
        \multicolumn{2}{c}{{\footnotesize $90.5 \pm 0.2$}} &
        \multicolumn{2}{c}{{\footnotesize $77.8 \pm 0.2$}} &
        \multicolumn{2}{c}{{\footnotesize $32.3 \pm 0.9$}} &
        \multicolumn{2}{c}{{\footnotesize $30.9 \pm 0.2$}}
        \\
        SSL4EO~\cite{ssl4eo} &
        \multicolumn{2}{c}{{\footnotesize $83.9 \pm 0.0$}} &
        \multicolumn{2}{c}{{\footnotesize $65.0 \pm 0.3$}} &
        \multicolumn{2}{c}{{\footnotesize $11.6 \pm 0.4$}} &
        \multicolumn{2}{c}{{\footnotesize $19.3 \pm 0.3$}} &
        \multicolumn{2}{c}{{\footnotesize $91.2 \pm 0.2$}} &
        \multicolumn{2}{c}{{\footnotesize $78.5 \pm 0.2$}} &
        \multicolumn{2}{c}{{\footnotesize $32.6 \pm 0.1$}} &
        \multicolumn{2}{c}{{\footnotesize $31.5 \pm 0.2$}}
        \\
        DOFA~\cite{dofa} &
        \multicolumn{2}{c}{{\footnotesize $83.1 \pm 0.1$}} &
        \multicolumn{2}{c}{{\footnotesize $63.1 \pm 0.5$}} &
        \multicolumn{2}{c}{{\footnotesize $13.5 \pm 0.5$}} &
        \multicolumn{2}{c}{{\footnotesize $17.6 \pm 0.5$}} &
        \multicolumn{2}{c}{{\footnotesize $90.7 \pm 0.1$}} &
        \multicolumn{2}{c}{{\footnotesize $77.3 \pm 0.3$}} &
        \multicolumn{2}{c}{{\footnotesize $\underline{34.8} \pm 0.9$}} &
        \multicolumn{2}{c}{{\footnotesize $29.9 \pm 0.3$}}
        \\
        \arrayrulecolor{gray} \midrule
        \rowcolor{lightblue} \textbf{SeCo-Eco (ours)} &
        \multicolumn{2}{c}{{\footnotesize $\textbf{84.8} \pm 0.2$}} &
        \multicolumn{2}{c}{{\footnotesize $\underline{65.6} \pm 0.2$}} &
        \multicolumn{2}{c}{{\footnotesize $\underline{14.8} \pm 0.6$}} &
        \multicolumn{2}{c}{{\footnotesize $\underline{19.9} \pm 0.2$}} &
        \multicolumn{2}{c}{{\footnotesize $\textbf{91.7} \pm 0.1$}} &
        \multicolumn{2}{c}{{\footnotesize $\underline{79.0} \pm 0.1$}} &
        \multicolumn{2}{c}{{\footnotesize $\textbf{35.7} \pm 0.4$}} &
        \multicolumn{2}{c}{{\footnotesize $\underline{32.4} \pm 0.2$}}
        \\
        \bottomrule
    \end{NiceTabular}
    \caption{
        Linear probing and K-Nearest Neighbor performance across multiple metrics for the EUForest task.
        \textbf{Best}, \underline{second best}.
    } 
    \label{table:all_metrics_euforest} 
\end{table*}
\definecolor{lightblue}{RGB}{227, 234, 250}
\begin{table*}[ht]
    \centering
    \small
    \setlength{\tabcolsep}{4pt} 
    \renewcommand{\arraystretch}{1.2} 
    \begin{NiceTabular}{@{}lcccccccccccccccc@{}}\\
        \toprule 
        \multirow{5}{*}{Model} &
        \multicolumn{16}{c}{TreeSatAI~\cite{ahlswede2022treesatai}} 
        \\
        \cmidrule(lr){2-17} &
        \multicolumn{4}{c}{Macro F1 $\uparrow$} &
        \multicolumn{4}{c}{Macro MAP $\uparrow$} &
        \multicolumn{4}{c}{Micro F1 $\uparrow$} &
        \multicolumn{4}{c}{Micro MAP $\uparrow$} 
        \\
        \cmidrule(lr){2-5} \cmidrule(lr){6-9} \cmidrule(lr){10-13} \cmidrule(lr){14-17} &
        \multicolumn{2}{c}{LP} &
        \multicolumn{2}{c}{5-NN} &
        \multicolumn{2}{c}{LP} &
        \multicolumn{2}{c}{5-NN} &
        \multicolumn{2}{c}{LP} &
        \multicolumn{2}{c}{5-NN} &
        \multicolumn{2}{c}{LP} &
        \multicolumn{2}{c}{5-NN}
        \\
        \midrule
        SeCo~\cite{seco} &
        \multicolumn{2}{c}{{\footnotesize $10.1 \pm 0.0$}} &
        \multicolumn{2}{c}{{\footnotesize $24.3$}} &
        \multicolumn{2}{c}{{\footnotesize $24.3 \pm 0.0$}} &
        \multicolumn{2}{c}{{\footnotesize $20.5$}} &
        \multicolumn{2}{c}{{\footnotesize $23.4 \pm 0.0$}} &
        \multicolumn{2}{c}{{\footnotesize $35.2$}} &
        \multicolumn{2}{c}{{\footnotesize $44.6 \pm 0.0$}} &
        \multicolumn{2}{c}{{\footnotesize $34.6$}}
        \\
        SatMAE~\cite{satmae} &
        \multicolumn{2}{c}{{\footnotesize $\textbf{21.0} \pm 0.1$}} &
        \multicolumn{2}{c}{{\footnotesize $\textbf{33.7}$}} &
        \multicolumn{2}{c}{{\footnotesize $\textbf{36.8} \pm 0.1$}} &
        \multicolumn{2}{c}{{\footnotesize $\textbf{35.8}$}} &
        \multicolumn{2}{c}{{\footnotesize $\textbf{46.8} \pm 0.3$}} &
        \multicolumn{2}{c}{{\footnotesize $\textbf{43.7}$}} &
        \multicolumn{2}{c}{{\footnotesize $\textbf{58.0} \pm 0.1$}} &
        \multicolumn{2}{c}{{\footnotesize $\textbf{52.3}$}}
        \\
        Satlas~\cite{saltas} &
        \multicolumn{2}{c}{{\footnotesize $17.8 \pm 0.0$}} &
        \multicolumn{2}{c}{{\footnotesize $30.1$}} &
        \multicolumn{2}{c}{{\footnotesize $32.4 \pm 0.0$}} &
        \multicolumn{2}{c}{{\footnotesize $27.9$}} &
        \multicolumn{2}{c}{{\footnotesize $42.9 \pm 0.0$}} &
        \multicolumn{2}{c}{{\footnotesize $40.8$}} &
        \multicolumn{2}{c}{{\footnotesize $54.2 \pm 0.0$}} &
        \multicolumn{2}{c}{{\footnotesize $45.4$}}
        \\
        Croma~\cite{croma} &
        \multicolumn{2}{c}{{\footnotesize $\underline{20.3} \pm 0.0$}} &
        \multicolumn{2}{c}{{\footnotesize $30.1$}} &
        \multicolumn{2}{c}{{\footnotesize $\underline{34.9} \pm 0.0$}} &
        \multicolumn{2}{c}{{\footnotesize $27.8$}} &
        \multicolumn{2}{c}{{\footnotesize $\underline{43.8} \pm 0.0$}} &
        \multicolumn{2}{c}{{\footnotesize $40.7$}} &
        \multicolumn{2}{c}{{\footnotesize $\underline{56.6} \pm 0.0$}} &
        \multicolumn{2}{c}{{\footnotesize $45.6$}}
        \\
        SSL4EO~\cite{ssl4eo} &
        \multicolumn{2}{c}{{\footnotesize $18.2 \pm 0.0$}} &
        \multicolumn{2}{c}{{\footnotesize $\underline{30.2}$}} &
        \multicolumn{2}{c}{{\footnotesize $33.1 \pm 0.0$}} &
        \multicolumn{2}{c}{{\footnotesize $28.4$}} &
        \multicolumn{2}{c}{{\footnotesize $42.3 \pm 0.0$}} &
        \multicolumn{2}{c}{{\footnotesize $\underline{40.9}$}} &
        \multicolumn{2}{c}{{\footnotesize $54.5 \pm 0.0$}} &
        \multicolumn{2}{c}{{\footnotesize $\underline{46.0}$}}
        \\
        DOFA~\cite{dofa} &
        \multicolumn{2}{c}{{\footnotesize $14.7 \pm 0.0$}} &
        \multicolumn{2}{c}{{\footnotesize $26.2$}} &
        \multicolumn{2}{c}{{\footnotesize $28.7 \pm 0.0$}} &
        \multicolumn{2}{c}{{\footnotesize $21.9$}} &
        \multicolumn{2}{c}{{\footnotesize $35.1 \pm 0.0$}} &
        \multicolumn{2}{c}{{\footnotesize $37.3$}} &
        \multicolumn{2}{c}{{\footnotesize $50.8 \pm 0.0$}} &
        \multicolumn{2}{c}{{\footnotesize $37.5$}}
        \\
        \arrayrulecolor{black} \midrule
        \rowcolor{lightblue} \textbf{SeCo-Eco (ours)} &
        \multicolumn{2}{c}{{\footnotesize $19.2 \pm 0.0$}} &
        \multicolumn{2}{c}{{\footnotesize $29.7$}} &
        \multicolumn{2}{c}{{\footnotesize $34.3 \pm 0.0$}} &
        \multicolumn{2}{c}{{\footnotesize $\underline{29.0}$}} &
        \multicolumn{2}{c}{{\footnotesize $42.7 \pm 0.0$}} &
        \multicolumn{2}{c}{{\footnotesize $40.6$}} &
        \multicolumn{2}{c}{{\footnotesize $54.8 \pm 0.0$}} &
        \multicolumn{2}{c}{{\footnotesize $45.7$}}
        \\
        \bottomrule
    \end{NiceTabular}
    \caption{
        Linear probing and K-Nearest Neighbor performance across multiple metrics for the TreeSatAI task.
        Due to the fixed splits, no standard deviation can be reported for K-Nearest Neighbor probing.
        \textbf{Best}, \underline{second best}.
    }
    \label{table:all_metrics_treesatai} 
\end{table*}

\definecolor{lightblue}{RGB}{227, 234, 250}
\begin{table*}[h]
    \centering
    \small
    \setlength{\tabcolsep}{4pt} 
    \renewcommand{\arraystretch}{1.2} 
    \begin{NiceTabular}{@{}lcccccccccccccccccccc@{}}\\
        \toprule 
        \multirow{5}{*}{Model} &
        \multicolumn{20}{c}{Biomes~\cite{biomes}} 
        \\
        \cmidrule(lr){2-21} &
        \multicolumn{4}{c}{Macro Acc $\uparrow$}  &
        \multicolumn{4}{c}{Macro AUROC $\uparrow$}  &
        \multicolumn{4}{c}{Macro F1 $\uparrow$}  &
        \multicolumn{4}{c}{Micro Acc $\uparrow$}  &
        \multicolumn{4}{c}{Micro F1 $\uparrow$} 
        \\
        \cmidrule(lr){2-5} \cmidrule(lr){6-9} \cmidrule(lr){10-13} \cmidrule(lr){14-17} \cmidrule(lr){18-21} &
        \multicolumn{2}{c}{LP}  &
        \multicolumn{2}{c}{10-NN} &
        \multicolumn{2}{c}{LP}  &
        \multicolumn{2}{c}{10-NN} &
        \multicolumn{2}{c}{LP}  &
        \multicolumn{2}{c}{10-NN} &
        \multicolumn{2}{c}{LP}  &
        \multicolumn{2}{c}{10-NN} &
        \multicolumn{2}{c}{LP}  &
        \multicolumn{2}{c}{10-NN}
        \\
        \midrule
        SeCo~\cite{seco} &
        \multicolumn{2}{c}{{\footnotesize $40.0 \pm 0.4$}} &
        \multicolumn{2}{c}{{\footnotesize $35.4 \pm 0.7$}} &
        \multicolumn{2}{c}{{\footnotesize $91.2 \pm 0.6$}} &
        \multicolumn{2}{c}{{\footnotesize $79.8 \pm 1.0$}} &
        \multicolumn{2}{c}{{\footnotesize $41.6 \pm 0.5$}} &
        \multicolumn{2}{c}{{\footnotesize $36.9 \pm 1.0$}} &
        \multicolumn{2}{c}{{\footnotesize $62.7 \pm 0.5$}} &
        \multicolumn{2}{c}{{\footnotesize $59.2 \pm 0.5$}} &
        \multicolumn{2}{c}{{\footnotesize $62.7 \pm 0.5$}} &
        \multicolumn{2}{c}{{\footnotesize $59.2 \pm 0.5$}}
        \\
        SatMAE~\cite{satmae} &
        \multicolumn{2}{c}{{\footnotesize $49.9 \pm 1.0$}} &
        \multicolumn{2}{c}{{\footnotesize $46.1 \pm 0.5$}} &
        \multicolumn{2}{c}{{\footnotesize $93.7 \pm 0.4$}} &
        \multicolumn{2}{c}{{\footnotesize $88.8 \pm 0.4$}} &
        \multicolumn{2}{c}{{\footnotesize $51.4 \pm 1.1$}} &
        \multicolumn{2}{c}{{\footnotesize $47.8 \pm 0.7$}} &
        \multicolumn{2}{c}{{\footnotesize $69.0 \pm 0.5$}} &
        \multicolumn{2}{c}{{\footnotesize $66.7 \pm 0.6$}} &
        \multicolumn{2}{c}{{\footnotesize $69.0 \pm 0.5$}} &
        \multicolumn{2}{c}{{\footnotesize $66.7 \pm 0.6$}}
        \\
        Satlas~\cite{saltas} &
        \multicolumn{2}{c}{{\footnotesize $47.1 \pm 1.4$}} &
        \multicolumn{2}{c}{{\footnotesize $45.9 \pm 0.7$}} &
        \multicolumn{2}{c}{{\footnotesize $92.8 \pm 0.5$}} &
        \multicolumn{2}{c}{{\footnotesize $88.4 \pm 0.4$}} &
        \multicolumn{2}{c}{{\footnotesize $48.3 \pm 1.6$}} &
        \multicolumn{2}{c}{{\footnotesize $47.6 \pm 0.9$}} &
        \multicolumn{2}{c}{{\footnotesize $65.6 \pm 0.8$}} &
        \multicolumn{2}{c}{{\footnotesize $65.1 \pm 0.5$}} &
        \multicolumn{2}{c}{{\footnotesize $65.6 \pm 0.8$}} &
        \multicolumn{2}{c}{{\footnotesize $65.1 \pm 0.5$}}
        \\
        Croma~\cite{croma} &
        \multicolumn{2}{c}{{\footnotesize $46.2 \pm 1.8$}} &
        \multicolumn{2}{c}{{\footnotesize $41.2 \pm 0.5$}} &
        \multicolumn{2}{c}{{\footnotesize $92.2 \pm 0.4$}} &
        \multicolumn{2}{c}{{\footnotesize $85.7 \pm 0.6$}} &
        \multicolumn{2}{c}{{\footnotesize $47.2 \pm 1.4$}} &
        \multicolumn{2}{c}{{\footnotesize $42.2 \pm 0.6$}} &
        \multicolumn{2}{c}{{\footnotesize $65.7 \pm 0.7$}} &
        \multicolumn{2}{c}{{\footnotesize $61.7 \pm 0.3$}} &
        \multicolumn{2}{c}{{\footnotesize $65.7 \pm 0.7$}} &
        \multicolumn{2}{c}{{\footnotesize $61.7 \pm 0.3$}}
        \\
        SSL4EO~\cite{ssl4eo} &
        \multicolumn{2}{c}{{\footnotesize $\underline{51.3} \pm 0.9$}} &
        \multicolumn{2}{c}{{\footnotesize $\underline{48.2} \pm 0.5$}} &
        \multicolumn{2}{c}{{\footnotesize $\underline{94.3} \pm 0.6$}} &
        \multicolumn{2}{c}{{\footnotesize $\underline{89.6} \pm 0.8$}} &
        \multicolumn{2}{c}{{\footnotesize $\underline{53.4} \pm 1.0$}} &
        \multicolumn{2}{c}{{\footnotesize $\underline{49.7} \pm 0.5$}} &
        \multicolumn{2}{c}{{\footnotesize $\underline{70.4} \pm 0.5$}} &
        \multicolumn{2}{c}{{\footnotesize $\underline{67.6} \pm 0.6$}} &
        \multicolumn{2}{c}{{\footnotesize $\underline{70.4} \pm 0.5$}} &
        \multicolumn{2}{c}{{\footnotesize $\underline{67.6} \pm 0.6$}}
        \\
        DOFA~\cite{dofa} &
        \multicolumn{2}{c}{{\footnotesize $48.1 \pm 1.4$}} &
        \multicolumn{2}{c}{{\footnotesize $41.8 \pm 0.4$}} &
        \multicolumn{2}{c}{{\footnotesize $92.9 \pm 0.3$}} &
        \multicolumn{2}{c}{{\footnotesize $85.7 \pm 0.6$}} &
        \multicolumn{2}{c}{{\footnotesize $49.7 \pm 1.3$}} &
        \multicolumn{2}{c}{{\footnotesize $43.0 \pm 0.5$}} &
        \multicolumn{2}{c}{{\footnotesize $66.4 \pm 0.6$}} &
        \multicolumn{2}{c}{{\footnotesize $61.8 \pm 0.5$}} &
        \multicolumn{2}{c}{{\footnotesize $66.4 \pm 0.6$}} &
        \multicolumn{2}{c}{{\footnotesize $61.8 \pm 0.5$}}
        \\
        \arrayrulecolor{gray} \midrule
        \rowcolor{lightblue} \textbf{SeCo-Eco (ours)} &
        \multicolumn{2}{c}{{\footnotesize $\textbf{53.9} \pm 0.7$}} &
        \multicolumn{2}{c}{{\footnotesize $\textbf{49.3} \pm 0.7$}} &
        \multicolumn{2}{c}{{\footnotesize $\textbf{95.5} \pm 0.4$}} &
        \multicolumn{2}{c}{{\footnotesize $\textbf{90.0} \pm 0.7$}} &
        \multicolumn{2}{c}{{\footnotesize $\textbf{56.1} \pm 0.7$}} &
        \multicolumn{2}{c}{{\footnotesize $\textbf{51.2} \pm 0.9$}} &
        \multicolumn{2}{c}{{\footnotesize $\textbf{72.9} \pm 0.5$}} &
        \multicolumn{2}{c}{{\footnotesize $\textbf{69.4} \pm 0.4$}} &
        \multicolumn{2}{c}{{\footnotesize $\textbf{72.9} \pm 0.5$}} &
        \multicolumn{2}{c}{{\footnotesize $\textbf{69.4} \pm 0.4$}}
        \\
        \bottomrule
    \end{NiceTabular}
    \caption{
        Linear probing and K-Nearest Neighbor performance across multiple metrics for the biomes classification task.
        \textbf{Best}, \underline{second best}.
    } 
    \label{table:all_metrics_biomes} 
\end{table*}
\definecolor{lightblue}{RGB}{227, 234, 250}
\begin{table*}[h]
    \centering
    \small
    \setlength{\tabcolsep}{4pt} 
    \renewcommand{\arraystretch}{1.2} 
    \begin{NiceTabular}{@{}lcccccccccccccccccccc@{}}\\
        \toprule 
        \multirow{5}{*}{Model} &
        \multicolumn{20}{c}{CAVM~\cite{cavm}} 
        \\
        \cmidrule(lr){2-21} &
        \multicolumn{4}{c}{Macro Acc $\uparrow$}  &
        \multicolumn{4}{c}{Macro AUROC $\uparrow$}  &
        \multicolumn{4}{c}{Macro F1 $\uparrow$}  &
        \multicolumn{4}{c}{Micro Acc $\uparrow$}  &
        \multicolumn{4}{c}{Micro F1 $\uparrow$} 
        \\
        \cmidrule(lr){2-5} \cmidrule(lr){6-9} \cmidrule(lr){10-13} \cmidrule(lr){14-17} \cmidrule(lr){18-21} &
        \multicolumn{2}{c}{LP} &
        \multicolumn{2}{c}{20-NN} &
        \multicolumn{2}{c}{LP} &
        \multicolumn{2}{c}{20-NN} &
        \multicolumn{2}{c}{LP} &
        \multicolumn{2}{c}{20-NN} &
        \multicolumn{2}{c}{LP} &
        \multicolumn{2}{c}{20-NN} &
        \multicolumn{2}{c}{LP} &
        \multicolumn{2}{c}{20-NN}
        \\
        \midrule
        SeCo~\cite{seco} &
        \multicolumn{2}{c}{{\footnotesize $53.2 \pm 0.6$}} &
        \multicolumn{2}{c}{{\footnotesize $50.3 \pm 0.6$}} &
        \multicolumn{2}{c}{{\footnotesize $87.3 \pm 0.3$}} &
        \multicolumn{2}{c}{{\footnotesize $85.6 \pm 0.3$}} &
        \multicolumn{2}{c}{{\footnotesize $54.5 \pm 0.7$}} &
        \multicolumn{2}{c}{{\footnotesize $52.1 \pm 0.7$}} &
        \multicolumn{2}{c}{{\footnotesize $61.4 \pm 0.6$}} &
        \multicolumn{2}{c}{{\footnotesize $60.6 \pm 0.5$}} &
        \multicolumn{2}{c}{{\footnotesize $61.4 \pm 0.6$}} &
        \multicolumn{2}{c}{{\footnotesize $60.6 \pm 0.5$}}
        \\
        SatMAE~\cite{satmae} &
        \multicolumn{2}{c}{{\footnotesize $55.2 \pm 1.6$}} &
        \multicolumn{2}{c}{{\footnotesize $54.0 \pm 0.6$}} &
        \multicolumn{2}{c}{{\footnotesize $88.3 \pm 0.3$}} &
        \multicolumn{2}{c}{{\footnotesize $87.9 \pm 0.3$}} &
        \multicolumn{2}{c}{{\footnotesize $56.4 \pm 1.5$}} &
        \multicolumn{2}{c}{{\footnotesize $55.8 \pm 0.7$}} &
        \multicolumn{2}{c}{{\footnotesize $63.0 \pm 0.5$}} &
        \multicolumn{2}{c}{{\footnotesize $63.5 \pm 0.5$}} &
        \multicolumn{2}{c}{{\footnotesize $63.0 \pm 0.5$}} &
        \multicolumn{2}{c}{{\footnotesize $63.5 \pm 0.5$}}
        \\
        Satlas~\cite{saltas} &
        \multicolumn{2}{c}{{\footnotesize $52.7 \pm 2.1$}} &
        \multicolumn{2}{c}{{\footnotesize $51.5 \pm 0.4$}} &
        \multicolumn{2}{c}{{\footnotesize $87.6 \pm 0.3$}} &
        \multicolumn{2}{c}{{\footnotesize $86.6 \pm 0.3$}} &
        \multicolumn{2}{c}{{\footnotesize $53.8 \pm 2.0$}} &
        \multicolumn{2}{c}{{\footnotesize $53.2 \pm 0.5$}} &
        \multicolumn{2}{c}{{\footnotesize $61.2 \pm 0.5$}} &
        \multicolumn{2}{c}{{\footnotesize $61.2 \pm 0.5$}} &
        \multicolumn{2}{c}{{\footnotesize $61.2 \pm 0.5$}} &
        \multicolumn{2}{c}{{\footnotesize $61.2 \pm 0.5$}}
        \\
        Croma~\cite{croma} &
        \multicolumn{2}{c}{{\footnotesize $52.7 \pm 1.3$}} &
        \multicolumn{2}{c}{{\footnotesize $50.1 \pm 0.7$}} &
        \multicolumn{2}{c}{{\footnotesize $87.4 \pm 0.3$}} &
        \multicolumn{2}{c}{{\footnotesize $85.6 \pm 0.4$}} &
        \multicolumn{2}{c}{{\footnotesize $53.7 \pm 1.2$}} &
        \multicolumn{2}{c}{{\footnotesize $51.6 \pm 0.8$}} &
        \multicolumn{2}{c}{{\footnotesize $61.0 \pm 0.7$}} &
        \multicolumn{2}{c}{{\footnotesize $60.3 \pm 0.6$}} &
        \multicolumn{2}{c}{{\footnotesize $61.0 \pm 0.7$}} &
        \multicolumn{2}{c}{{\footnotesize $60.3 \pm 0.6$}}
        \\
        SSL4EO~\cite{ssl4eo} &
        \multicolumn{2}{c}{{\footnotesize $\underline{56.0} \pm 0.5$}} &
        \multicolumn{2}{c}{{\footnotesize $\underline{55.0} \pm 0.6$}} &
        \multicolumn{2}{c}{{\footnotesize $\underline{88.9} \pm 0.3$}} &
        \multicolumn{2}{c}{{\footnotesize $\underline{88.2} \pm 0.3$}} &
        \multicolumn{2}{c}{{\footnotesize $\underline{57.5} \pm 0.6$}} &
        \multicolumn{2}{c}{{\footnotesize $\underline{56.9} \pm 0.7$}} &
        \multicolumn{2}{c}{{\footnotesize $\underline{63.7} \pm 0.6$}} &
        \multicolumn{2}{c}{{\footnotesize $\underline{63.7} \pm 0.5$}} &
        \multicolumn{2}{c}{{\footnotesize $\underline{63.7} \pm 0.6$}} &
        \multicolumn{2}{c}{{\footnotesize $\underline{63.7} \pm 0.5$}}
        \\
        DOFA~\cite{dofa} &
        \multicolumn{2}{c}{{\footnotesize $55.3 \pm 1.8$}} &
        \multicolumn{2}{c}{{\footnotesize $51.7 \pm 0.5$}} &
        \multicolumn{2}{c}{{\footnotesize $88.2 \pm 0.4$}} &
        \multicolumn{2}{c}{{\footnotesize $87.0 \pm 0.3$}} &
        \multicolumn{2}{c}{{\footnotesize $56.5 \pm 1.6$}} &
        \multicolumn{2}{c}{{\footnotesize $53.6 \pm 0.6$}} &
        \multicolumn{2}{c}{{\footnotesize $62.4 \pm 0.8$}} &
        \multicolumn{2}{c}{{\footnotesize $62.2 \pm 0.4$}} &
        \multicolumn{2}{c}{{\footnotesize $62.4 \pm 0.8$}} &
        \multicolumn{2}{c}{{\footnotesize $62.2 \pm 0.4$}}
        \\
        \arrayrulecolor{gray} \midrule
        \rowcolor{lightblue} \textbf{SeCo-Eco (ours)} &
        \multicolumn{2}{c}{{\footnotesize $\textbf{58.1} \pm 1.2$}} &
        \multicolumn{2}{c}{{\footnotesize $\textbf{58.0} \pm 0.7$}} &
        \multicolumn{2}{c}{{\footnotesize $\textbf{89.9} \pm 0.3$}} &
        \multicolumn{2}{c}{{\footnotesize $\textbf{89.2} \pm 0.4$}} &
        \multicolumn{2}{c}{{\footnotesize $\textbf{59.4} \pm 1.0$}} &
        \multicolumn{2}{c}{{\footnotesize $\textbf{59.5} \pm 0.8$}} &
        \multicolumn{2}{c}{{\footnotesize $\textbf{65.3} \pm 0.5$}} &
        \multicolumn{2}{c}{{\footnotesize $\textbf{65.6} \pm 0.6$}} &
        \multicolumn{2}{c}{{\footnotesize $\textbf{65.3} \pm 0.5$}} &
        \multicolumn{2}{c}{{\footnotesize $\textbf{65.6} \pm 0.6$}}
        \\
        \bottomrule
        \end{NiceTabular}
    \caption{Linear probing and K-Nearest Neighbor performance across multiple metrics for the CAVM classification task. \textbf{Best}, \underline{second best}.} 
    \label{table:all_metrics_arctic} 
\end{table*}

 \definecolor{lightblue}{RGB}{227, 234, 250}
\begin{table*}[ht]
    \centering
    \small
    \setlength{\tabcolsep}{4pt} 
    \renewcommand{\arraystretch}{1.2} 
    \begin{NiceTabular}{@{}lcccccccccccc@{}}\\
        \toprule 
        \multirow{3}{*}{Model} &
        \multicolumn{12}{c}{BioMassters~\cite{biomassters}} 
        \\
        \cmidrule(lr){2-13} &
        \multicolumn{4}{c}{Mean R\textsuperscript{2} $\uparrow$} &
        \multicolumn{4}{c}{Mean MAE $\downarrow$} &
        \multicolumn{4}{c}{Mean RMSE $\downarrow$} 
        \\
        \cmidrule(lr){2-5} \cmidrule(lr){6-9} \cmidrule(lr){10-13} &
        \multicolumn{2}{c}{LP} &
        \multicolumn{2}{c}{1-NN} &
        \multicolumn{2}{c}{LP} &
        \multicolumn{2}{c}{1-NN} &
        \multicolumn{2}{c}{LP} &
        \multicolumn{2}{c}{1-NN}
        \\
        \midrule
        SeCo~\cite{seco} &
        \multicolumn{2}{c}{{\footnotesize $51.3 \pm 0.0$}} &
        \multicolumn{2}{c}{{\footnotesize$-19.2$}} &
        \multicolumn{2}{c}{{\footnotesize $3.9 \pm 0.0$}} &
        \multicolumn{2}{c}{{\footnotesize$\underline{7.0}$}} &
        \multicolumn{2}{c}{{\footnotesize $5.8 \pm 0.0$}} &
        \multicolumn{2}{c}{{\footnotesize$\underline{11.0}$}}
        \\
        SatMAE~\cite{satmae} &
        \multicolumn{2}{c}{{\footnotesize $59.5 \pm 0.6$}} &
        \multicolumn{2}{c}{{\footnotesize$-18.0$}} &
        \multicolumn{2}{c}{{\footnotesize $3.6 \pm 0.0$}} &
        \multicolumn{2}{c}{{\footnotesize$\underline{7.0} $}} &
        \multicolumn{2}{c}{{\footnotesize $5.3 \pm 0.0$}} &
        \multicolumn{2}{c}{{\footnotesize$\underline{11.0}$}}
        \\
        Satlas~\cite{saltas} &
        \multicolumn{2}{c}{{\footnotesize $62.5 \pm 0.9$}} &
        \multicolumn{2}{c}{{\footnotesize$-17.8$}} &
        \multicolumn{2}{c}{{\footnotesize $3.3 \pm 0.1$}} &
        \multicolumn{2}{c}{{\footnotesize$\underline{7.0}$}} &
        \multicolumn{2}{c}{{\footnotesize $4.9 \pm 0.1$}} &
        \multicolumn{2}{c}{{\footnotesize$\underline{11.0}$}}
        \\
        Croma~\cite{croma} &
        \multicolumn{2}{c}{{\footnotesize $58.5 \pm 0.2$}} &
        \multicolumn{2}{c}{{\footnotesize$-18.1$}} &
        \multicolumn{2}{c}{{\footnotesize $3.5 \pm 0.0$}} &
        \multicolumn{2}{c}{{\footnotesize$\underline{7.0} $}} &
        \multicolumn{2}{c}{{\footnotesize $5.3 \pm 0.0$}} &
        \multicolumn{2}{c}{{\footnotesize$\underline{11.0}$}}
        \\
        SSL4EO~\cite{ssl4eo} &
        \multicolumn{2}{c}{{\footnotesize $\underline{71.4} \pm 0.0$}} &
        \multicolumn{2}{c}{{\footnotesize$\underline{-16.8}$}} &
        \multicolumn{2}{c}{{\footnotesize $\underline{2.8} \pm 0.0$}} &
        \multicolumn{2}{c}{{\footnotesize$\textbf{6.9} $}} &
        \multicolumn{2}{c}{{\footnotesize $\underline{4.2} \pm 0.0$}} &
        \multicolumn{2}{c}{{\footnotesize$\textbf{10.9}$}}
        \\
        DOFA~\cite{dofa} &
        \multicolumn{2}{c}{{\footnotesize $63.1 \pm 0.4$}} &
        \multicolumn{2}{c}{{\footnotesize$-18.3$}} &
        \multicolumn{2}{c}{{\footnotesize $3.2 \pm 0.0$}} &
        \multicolumn{2}{c}{{\footnotesize$\underline{7.0} $}} &
        \multicolumn{2}{c}{{\footnotesize $4.8 \pm 0.0$}} &
        \multicolumn{2}{c}{{\footnotesize$\underline{11.0}$}}
        \\
        \arrayrulecolor{black} \midrule
        \rowcolor{lightblue} \textbf{SeCo-Eco (ours)} &
        \multicolumn{2}{c}{{\footnotesize $\textbf{75.2} \pm 0.1$}} &
        \multicolumn{2}{c}{{\footnotesize$\textbf{-16.3}$}} &
        \multicolumn{2}{c}{{\footnotesize $\textbf{2.5} \pm 0.0$}} &
        \multicolumn{2}{c}{{\footnotesize$\textbf{6.9}$}} &
        \multicolumn{2}{c}{{\footnotesize $\textbf{3.8} \pm 0.0$}} &
        \multicolumn{2}{c}{{\footnotesize$\textbf{10.9}$}}
        \\
        \bottomrule
    \end{NiceTabular}
    \caption{
        Linear probing and K-Nearest Neighbor performance across multiple metrics for the BioMassters task.
        Due to the fixed splits, no standard deviation can be reported for K-Nearest Neighbor probing.
        \textbf{Best}, \underline{second best}.
    }
    \label{tab:all_metrics_biomassters} 
\end{table*}
\definecolor{lightblue}{RGB}{227, 234, 250}
\begin{table*}[h]
    \centering
    \small
    \setlength{\tabcolsep}{4pt} 
    \renewcommand{\arraystretch}{1.2} 
    \begin{NiceTabular}{@{}lcccccccccccccccc@{}}\\
        \toprule 
        \multirow{5}{*}{Model} &
        \multicolumn{16}{c}{CHELSA Climate~\cite{chelsa} - Temperature \& Precipitation} 
        \\
        \cmidrule(lr){2-17} &
        \multicolumn{4}{c}{Temp MAE $\downarrow$} &
        \multicolumn{4}{c}{Temp R$^2$ $\uparrow$} &
        \multicolumn{4}{c}{Prec MAE $\downarrow$} &
        \multicolumn{4}{c}{Prec R$^2$ $\uparrow$} 
        \\
        \cmidrule(lr){2-5} \cmidrule(lr){6-9} \cmidrule(lr){10-13} \cmidrule(lr){14-17} &
        \multicolumn{2}{c}{LP} &
        \multicolumn{2}{c}{10-NN} &
        \multicolumn{2}{c}{LP} &
        \multicolumn{2}{c}{10-NN} &
        \multicolumn{2}{c}{LP} &
        \multicolumn{2}{c}{10-NN} &
        \multicolumn{2}{c}{LP} &
        \multicolumn{2}{c}{10-NN}
        \\
        \midrule
        SeCo~\cite{seco} &
        \multicolumn{2}{c}{{\footnotesize $572.3 \pm 1.1$}} &
        \multicolumn{2}{c}{{\footnotesize $547.8 \pm 1.7$}} &
        \multicolumn{2}{c}{{\footnotesize $63.1 \pm 0.3$}} &
        \multicolumn{2}{c}{{\footnotesize $61.3 \pm 0.3$}} &
        \multicolumn{2}{c}{{\footnotesize $33380.8 \pm 291.5$}} &
        \multicolumn{2}{c}{{\footnotesize $30725.5 \pm 171.8$}} &
        \multicolumn{2}{c}{{\footnotesize $60.3 \pm 0.7$}} &
        \multicolumn{2}{c}{{\footnotesize $60.7 \pm 0.8$}}
        \\
        SatMAE~\cite{satmae} &
        \multicolumn{2}{c}{{\footnotesize $\underline{482.0} \pm 2.3$}} &
        \multicolumn{2}{c}{{\footnotesize $411.4 \pm 1.2$}} &
        \multicolumn{2}{c}{{\footnotesize $\underline{74.4} \pm 0.2$}} &
        \multicolumn{2}{c}{{\footnotesize $\underline{76.1} \pm 0.2$}} &
        \multicolumn{2}{c}{{\footnotesize $30999.5 \pm 314.9$}} &
        \multicolumn{2}{c}{{\footnotesize $\underline{27087.1} \pm 135.4$}} &
        \multicolumn{2}{c}{{\footnotesize $65.2 \pm 0.4$}} &
        \multicolumn{2}{c}{{\footnotesize $\underline{67.1} \pm 0.5$}}
        \\
        Satlas~\cite{saltas} &
        \multicolumn{2}{c}{{\footnotesize $595.1 \pm 3.4$}} &
        \multicolumn{2}{c}{{\footnotesize $474.7 \pm 3.6$}} &
        \multicolumn{2}{c}{{\footnotesize $62.1 \pm 0.4$}} &
        \multicolumn{2}{c}{{\footnotesize $69.4 \pm 0.7$}} &
        \multicolumn{2}{c}{{\footnotesize $36698.8 \pm 685.3$}} &
        \multicolumn{2}{c}{{\footnotesize $29535.8 \pm 95.1$}} &
        \multicolumn{2}{c}{{\footnotesize $55.9 \pm 1.0$}} &
        \multicolumn{2}{c}{{\footnotesize $62.4 \pm 0.7$}}
        \\
        Croma~\cite{croma} &
        \multicolumn{2}{c}{{\footnotesize $511.5 \pm 2.5$}} &
        \multicolumn{2}{c}{{\footnotesize $505.5 \pm 1.6$}} &
        \multicolumn{2}{c}{{\footnotesize $71.1 \pm 0.2$}} &
        \multicolumn{2}{c}{{\footnotesize $66.4 \pm 0.2$}} &
        \multicolumn{2}{c}{{\footnotesize $32887.8 \pm 350.8$}} &
        \multicolumn{2}{c}{{\footnotesize $30974.2 \pm 96.6$}} &
        \multicolumn{2}{c}{{\footnotesize $61.4 \pm 0.6$}} &
        \multicolumn{2}{c}{{\footnotesize $60.3 \pm 0.4$}}
        \\
        SSL4EO~\cite{ssl4eo} &
        \multicolumn{2}{c}{{\footnotesize $496.1 \pm 1.1$}} &
        \multicolumn{2}{c}{{\footnotesize $\underline{410.7} \pm 0.8$}} &
        \multicolumn{2}{c}{{\footnotesize $72.4 \pm 0.2$}} &
        \multicolumn{2}{c}{{\footnotesize $75.8 \pm 0.3$}} &
        \multicolumn{2}{c}{{\footnotesize $\underline{30960.7} \pm 154.7$}} &
        \multicolumn{2}{c}{{\footnotesize $27989.7 \pm 148.3$}} &
        \multicolumn{2}{c}{{\footnotesize $\underline{65.5} \pm 0.4$}} &
        \multicolumn{2}{c}{{\footnotesize $65.4 \pm 0.4$}}
        \\
        DOFA~\cite{dofa} &
        \multicolumn{2}{c}{{\footnotesize $576.0 \pm 0.7$}} &
        \multicolumn{2}{c}{{\footnotesize $505.9 \pm 0.9$}} &
        \multicolumn{2}{c}{{\footnotesize $63.9 \pm 0.3$}} &
        \multicolumn{2}{c}{{\footnotesize $66.9 \pm 0.3$}} &
        \multicolumn{2}{c}{{\footnotesize $34860.1 \pm 297.0$}} &
        \multicolumn{2}{c}{{\footnotesize $30311.1 \pm 182.9$}} &
        \multicolumn{2}{c}{{\footnotesize $59.7 \pm 0.5$}} &
        \multicolumn{2}{c}{{\footnotesize $59.9 \pm 0.7$}}
        \\
        \arrayrulecolor{gray} \midrule
        \rowcolor{lightblue} \textbf{SeCo-Eco (ours)} &
        \multicolumn{2}{c}{{\footnotesize $\textbf{411.4} \pm 0.9$}} &
        \multicolumn{2}{c}{{\footnotesize $\textbf{364.8} \pm 0.7$}} &
        \multicolumn{2}{c}{{\footnotesize $\textbf{80.7} \pm 0.2$}} &
        \multicolumn{2}{c}{{\footnotesize $\textbf{80.5} \pm 0.2$}} &
        \multicolumn{2}{c}{{\footnotesize $\textbf{27695.5} \pm 74.8$}} &
        \multicolumn{2}{c}{{\footnotesize $\textbf{25946.7} \pm 72.6$}} &
        \multicolumn{2}{c}{{\footnotesize $\textbf{70.2} \pm 0.3$}} &
        \multicolumn{2}{c}{{\footnotesize $\textbf{69.5} \pm 0.4$}}
        \\
        \bottomrule
    \end{NiceTabular}
    \caption{
        Linear probing and K-Nearest Neighbor performance overview for the CHELSA Climate task.
        We break down the predictions for temperature and precipitation.
        \textbf{Best}, \underline{second best}.
    } 
    \label{table:all_metrics_tempprec} 
\end{table*}
\definecolor{lightblue}{RGB}{227, 234, 250}
\begin{table*}[h]
    \centering
    \small
    \setlength{\tabcolsep}{4pt} 
    \renewcommand{\arraystretch}{1.2} 
    \begin{NiceTabular}{@{}lcccccccccccccccc@{}}\\
        \toprule 
        \multirow{5}{*}{Model}  &
        \multicolumn{16}{c}{CHELSA Climate~\cite{chelsa} - Evapotranspiration \& Site Water Balance} 
        \\
        \cmidrule(lr){2-17} &
        \multicolumn{4}{c}{Evap MAE $\downarrow$} &
        \multicolumn{4}{c}{Evap R$^2$ $\uparrow$} &
        \multicolumn{4}{c}{Swb MAE $\downarrow$} &
        \multicolumn{4}{c}{Swb R$^2$ $\uparrow$} 
        \\
        \cmidrule(lr){2-5} \cmidrule(lr){6-9} \cmidrule(lr){10-13} \cmidrule(lr){14-17} &
        \multicolumn{2}{c}{LP} &
        \multicolumn{2}{c}{10-NN} &
        \multicolumn{2}{c}{LP} &
        \multicolumn{2}{c}{10-NN} &
        \multicolumn{2}{c}{LP} &
        \multicolumn{2}{c}{10-NN} &
        \multicolumn{2}{c}{LP} &
        \multicolumn{2}{c}{10-NN}
        \\
        \midrule
        SeCo~\cite{seco} &
        \multicolumn{2}{c}{{\footnotesize $2131.6 \pm 8.7$}} &
        \multicolumn{2}{c}{{\footnotesize $2068.0 \pm 9.8$}} &
        \multicolumn{2}{c}{{\footnotesize $68.9 \pm 0.2$}} &
        \multicolumn{2}{c}{{\footnotesize $67.1 \pm 0.1$}} &
        \multicolumn{2}{c}{{\footnotesize $24903.7 \pm 137.7$}} &
        \multicolumn{2}{c}{{\footnotesize $23878.3 \pm 143.0$}} &
        \multicolumn{2}{c}{{\footnotesize $80.9 \pm 0.2$}} &
        \multicolumn{2}{c}{{\footnotesize $80.5 \pm 0.2$}}
        \\
        SatMAE~\cite{satmae} &
        \multicolumn{2}{c}{{\footnotesize $\underline{1760.4} \pm 6.7$}} &
        \multicolumn{2}{c}{{\footnotesize $1564.7 \pm 5.4$}} &
        \multicolumn{2}{c}{{\footnotesize $\underline{79.2} \pm 0.2$}} &
        \multicolumn{2}{c}{{\footnotesize $80.0 \pm 0.2$}} &
        \multicolumn{2}{c}{{\footnotesize $20999.2 \pm 87.5$}} &
        \multicolumn{2}{c}{{\footnotesize $19055.6 \pm 66.2$}} &
        \multicolumn{2}{c}{{\footnotesize $86.9 \pm 0.1$}} &
        \multicolumn{2}{c}{{\footnotesize $87.7 \pm 0.1$}}
        \\
        Satlas~\cite{saltas} &
        \multicolumn{2}{c}{{\footnotesize $2093.3 \pm 6.4$}} &
        \multicolumn{2}{c}{{\footnotesize $1761.5 \pm 10.3$}} &
        \multicolumn{2}{c}{{\footnotesize $70.9 \pm 0.2$}} &
        \multicolumn{2}{c}{{\footnotesize $75.3 \pm 0.5$}} &
        \multicolumn{2}{c}{{\footnotesize $24115.2 \pm 131.3$}} &
        \multicolumn{2}{c}{{\footnotesize $20772.5 \pm 116.8$}} &
        \multicolumn{2}{c}{{\footnotesize $83.4 \pm 0.2$}} &
        \multicolumn{2}{c}{{\footnotesize $85.5 \pm 0.1$}}
        \\
        Croma~\cite{croma} &
        \multicolumn{2}{c}{{\footnotesize $1872.4 \pm 27.2$}} &
        \multicolumn{2}{c}{{\footnotesize $1882.2 \pm 5.6$}} &
        \multicolumn{2}{c}{{\footnotesize $76.2 \pm 0.5$}} &
        \multicolumn{2}{c}{{\footnotesize $73.2 \pm 0.2$}} &
        \multicolumn{2}{c}{{\footnotesize $23003.2 \pm 270.2$}} &
        \multicolumn{2}{c}{{\footnotesize $21593.9 \pm 79.0$}} &
        \multicolumn{2}{c}{{\footnotesize $84.6 \pm 0.4$}} &
        \multicolumn{2}{c}{{\footnotesize $84.7 \pm 0.2$}}
        \\
        SSL4EO~\cite{ssl4eo} &
        \multicolumn{2}{c}{{\footnotesize $1786.0 \pm 3.4$}} &
        \multicolumn{2}{c}{{\footnotesize $\underline{1522.8} \pm 3.0$}} &
        \multicolumn{2}{c}{{\footnotesize $78.6 \pm 0.2$}} &
        \multicolumn{2}{c}{{\footnotesize $\underline{81.1} \pm 0.2$}} &
        \multicolumn{2}{c}{{\footnotesize $\underline{20444.3} \pm 58.1$}} &
        \multicolumn{2}{c}{{\footnotesize $\underline{18155.6} \pm 42.4$}} &
        \multicolumn{2}{c}{{\footnotesize $\underline{87.7} \pm 0.1$}} &
        \multicolumn{2}{c}{{\footnotesize $\underline{88.9} \pm 0.1$}}
        \\
        DOFA~\cite{dofa} &
        \multicolumn{2}{c}{{\footnotesize $2086.1 \pm 3.5$}} &
        \multicolumn{2}{c}{{\footnotesize $1911.6 \pm 6.2$}} &
        \multicolumn{2}{c}{{\footnotesize $71.2 \pm 0.3$}} &
        \multicolumn{2}{c}{{\footnotesize $72.0 \pm 0.3$}} &
        \multicolumn{2}{c}{{\footnotesize $23943.5 \pm 38.3$}} &
        \multicolumn{2}{c}{{\footnotesize $22370.0 \pm 60.8$}} &
        \multicolumn{2}{c}{{\footnotesize $83.4 \pm 0.1$}} &
        \multicolumn{2}{c}{{\footnotesize $83.5 \pm 0.2$}}
        \\
        \arrayrulecolor{gray} \midrule
        \rowcolor{lightblue} \textbf{SeCo-Eco (ours)} &
        \multicolumn{2}{c}{{\footnotesize $\textbf{1537.6} \pm 4.2$}} &
        \multicolumn{2}{c}{{\footnotesize $\textbf{1391.2} \pm 3.3$}} &
        \multicolumn{2}{c}{{\footnotesize $\textbf{83.7} \pm 0.1$}} &
        \multicolumn{2}{c}{{\footnotesize $\textbf{83.9} \pm 0.2$}} &
        \multicolumn{2}{c}{{\footnotesize $\textbf{18567.4} \pm 90.2$}} &
        \multicolumn{2}{c}{{\footnotesize $\textbf{17257.4} \pm 50.6$}} &
        \multicolumn{2}{c}{{\footnotesize $\textbf{89.6} \pm 0.1$}} &
        \multicolumn{2}{c}{{\footnotesize $\textbf{89.9} \pm 0.1$}}
        \\
        \bottomrule
        \end{NiceTabular}
    \caption{Linear probing and K-Nearest Neighbor performance overview for the CHELSA Climate task. We break down the predictions for evapotranspiration and site water balance. \textbf{Best}, \underline{second best}.} 
    \label{table:all_metrics_evapswb} 
\end{table*}


\end{document}